%% file: main.tex
\documentclass[10pt,twocolumn,letterpaper]{article}
\usepackage[]{cvpr}
\usepackage{amsmath,amssymb,amsfonts,mathtools,bm}
\usepackage{booktabs,tabularx,multirow,array}
\newcolumntype{Y}{>{\centering\arraybackslash}X}
\usepackage{algorithm,algpseudocode}
\usepackage{graphicx}
\usepackage{tikz}
\usepackage{placeins}
\usetikzlibrary{arrows.meta,positioning,calc,fit,shapes.geometric}
\usepackage{xcolor,microtype,enumitem}
\usepackage{needspace}
\definecolor{cvprblue}{rgb}{0.21,0.49,0.74}
\usepackage[pagebackref,breaklinks,colorlinks,allcolors=cvprblue]{hyperref}
\usepackage[all]{hypcap}
\usepackage{threeparttable}

\usepackage{xr}
\algrenewcommand\algorithmicrequire{\textbf{Input:}}
\algrenewcommand\algorithmicensure{\textbf{Output:}}
\def\confName{CVPR}\def\confYear{2027}

\title{Dynamics-Guided Conditional MeanFlow from 
Regional Interaction to Internal Revision and Its Application to CT Imaging}
\author{
Md Shifatul Ahsan Apurba$^{1}$\thanks{Corresponding author.}\qquad
Md Selim$^{2}$\qquad
Jin Chen$^{1,3}$\\
$^{1}$Department of Biomedical Informatics and Data Science\\
$^{3}$Division of Nephrology, Department of Medicine\\
University of Alabama at Birmingham, Birmingham, AL, USA\\
$^{2}$Department of Computer Science, Florida Polytechnic University, Lakeland, FL, USA\\
{\ttfamily\small mapurba@uab.edu \quad mselim@floridapoly.edu \quad jinchen@uab.edu}
} % Automatically anonymized by cvpr.sty in review mode
\date{}

\usepackage{graphicx}
\usepackage[percent]{overpic}
\usepackage{tabularx}
\usepackage{array}
\begin{document}
\maketitle

\begin{abstract}
Conditional image enhancement requires spatially varying corrections, yet fixed-depth models allocate similar computational budgets regardless of where additional refinement is needed. Although one-step flow models reduce inference costs and iterative restoration methods introduce subsequent corrections, how predicted transformation dynamics can guide internal feature refinement remains underexplored. We introduce IRCMF, a dynamics-guided Conditional MeanFlow framework that combines regional interaction with recurrent internal revision. The decoder predicts signed interval velocity and a learned velocity-change cue. A soft-region graph aggregates contextual information for velocity estimation, while a dynamics-conditioned graph establishes regional relationships to guide recurrent bottleneck updates. The revised features are processed by the shared decoder to recompute velocity, establishing feedback between transformation dynamics and internal representations. An actor-critic policy controls regional updates and further revisions to balance enhancement quality and computational cost. We evaluate IRCMF on two computed tomography (CT) datasets, NLST for STANDARD-to-LUNG kernel enhancement and LDCT-and-Projection-data for simulated low-to-routine-dose enhancement. With nine decoder evaluations, IRCMF achieves slice-level PSNR of $36.24 \pm 3.49$ dB and SSIM of $0.97 \pm 0.01$ on NLST, and PSNR of $32.45 \pm 1.97$ dB and SSIM of $0.84 \pm 0.05$ on LDCT-and-Projection-data. The learned policy reduces average decoder evaluations to 4.42 and 3.45, respectively, lowering per-slice latency from 97.00 to 51.82 ms and from 97.12 to 41.95 ms, with PSNR reductions of only 0.20 and 0.06 dB.
\end{abstract}

\begin{figure}[t]
    \centering
    \includegraphics[width=\columnwidth]{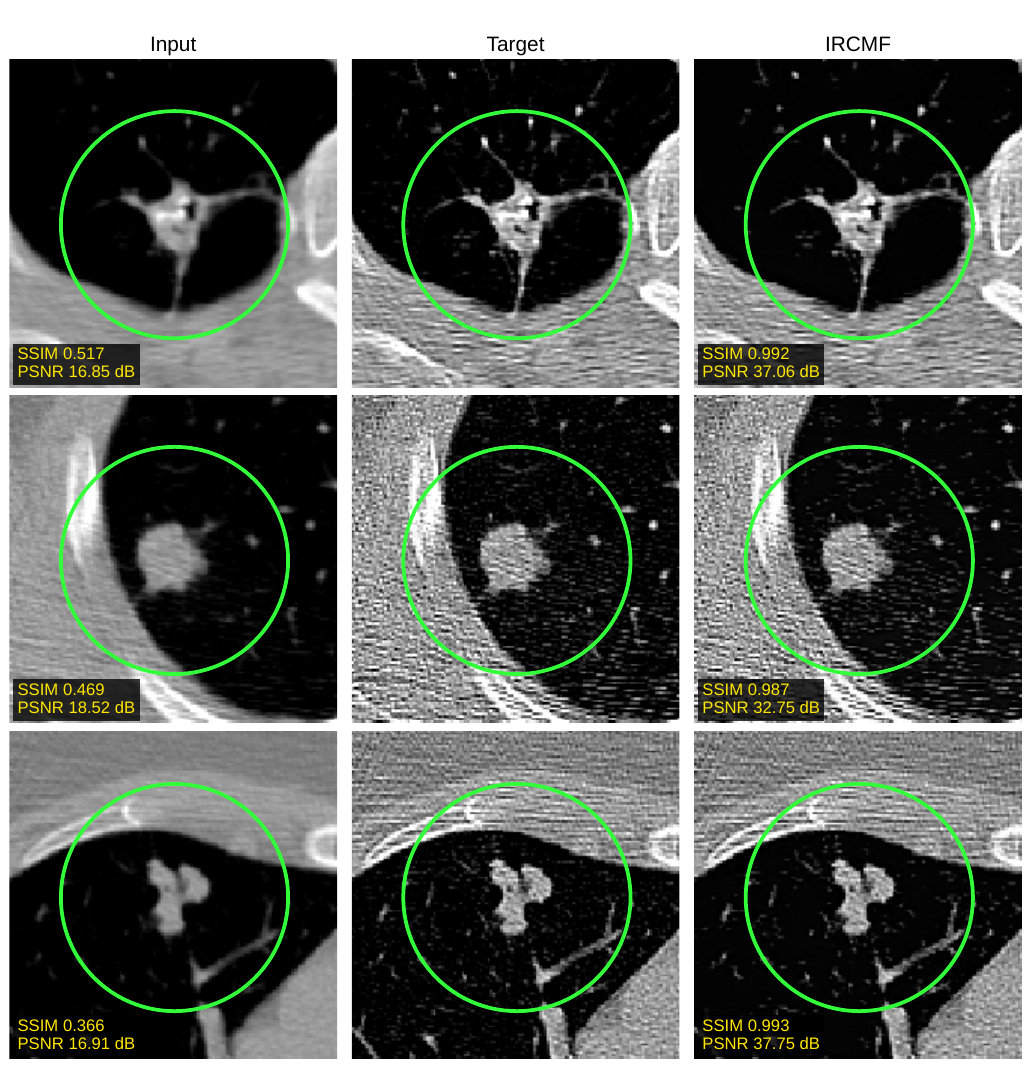}
        \caption{\textbf{Examples of tumor-region enhancement by IRCMF.} Each row shows an aligned crop from a different held-out NLST CT case, with Input, Target, and IRCMF (ours) shown from left to right. Equal-sized green circles locate the annotated tumors. The overlaid SSIM and PSNR compare Input and IRCMF with Target within the tumor masks.}
    \label{fig:tumor_roi_examples}
\end{figure}

\section{Introduction}
\label{sec:introduction}

Predicting an image transformation and deciding how to revise that prediction are distinct problems. In conditional image enhancement, required changes vary across an image: smooth regions may need modest intensity adjustment, whereas structural boundaries and textured regions may require more careful correction. Yet fixed-depth predictors use the same computational schedule even when the benefit of further refinement varies by location. Spatially adaptive restoration~\cite{purohit2021spair} and region-dependent sampling~\cite{liu2026ras,schusterbauer2026fastslow} address aspects of this inefficiency. For velocity-based predictors, the field describes the proposed transformation but does not directly indicate whether another evaluation will reduce error or how internal representations should change to produce a better estimate.

Flow Matching~\cite{lipman2023flowmatching} models generative transformations through learned velocity fields, while MeanFlow~\cite{geng2025meanflow} introduces interval-average velocity prediction for one-step generation. Improved Mean Flows~\cite{geng2025improved} examines its learning formulation, and Reversing Flow~\cite{qin2025resflow} develops deterministic restoration paths. Adaptive sampling~\cite{liu2026ras,zhu2026tap}, iterative latent restoration~\cite{chen2026rar}, and Region-Adaptive Conditional MeanFlow~\cite{apurba2026racmf} further explore allocating computation during refinement. These developments motivate using predicted dynamics to guide internal revision. Signed velocity describes the direction and magnitude of image changes, while a learned change cue provides complementary information. Although neither directly estimates error, both can inform regional interactions and guide revisions of the hidden features responsible for the \emph{next} velocity prediction, rather than only controlling image-space corrections or sampling schedules.

We introduce Interaction and Revision based Conditional MeanFlow (IRCMF), a dynamics-guided framework for regional interaction and internal feature revision within a conditional MeanFlow-derived U-Net~\cite{ronneberger2015unet}. A decoder-side soft-region graph pools feature and dynamic information and exchanges regional messages to contextualize signed interval-velocity prediction. The predicted velocity and learned velocity-change cue guide a second soft-region graph on bottleneck features, providing context for recurrent, spatially gated updates. The shared decoder recomputes velocity from the revised features, coupling regional reasoning and velocity re-estimation through internal feedback. A learned actor-critic policy~\cite{konda1999actorcritic} selects regions for continued updates, balancing full-image error reduction against decoder-call cost during training. At inference, decisions use only the observed input and model predictions. Regional gates control feature revision; each additional decoder evaluation remains dense and is skipped when no region continues.

Our contributions are threefold. (i) We introduce a dynamics-conditioned decoder that uses soft-region graph communication to refine signed interval-velocity predictions;
(ii) we develop graph-guided internal revision, where predicted velocity and a learned velocity-change cue guide recurrent bottleneck updates before the shared decoder recomputes velocity; and
(iii) we learn region-wise continuation with an actor-critic policy that balances full-image error reduction against dense decoder evaluations, and characterize the resulting quality-computation trade-off on two CT enhancement settings.

We study this formulation on two computed tomography (CT) datasets. (i) the National Lung Screening Trial (NLST)~\cite{nlst2013} for STANDARD-to-LUNG kernel enhancement and (ii) LDCT-and-Projection-data~\cite{mccollough2020ldct,moen2021ldct} for simulated low-to-routine-dose enhancement. With nine decoder evaluations, the framework achieves slice-level PSNR of $36.24$ dB and SSIM of $0.97$ on NLST, and PSNR of $32.45$ dB and SSIM of $0.84$ on LDCT-and-Projection-data. The learned policy reduces mean decoder evaluations to $4.42$ and $3.45$, respectively, and per-slice latency from $97.00$ to $51.82$ ms and from $97.12$ to $41.95$ ms, with corresponding PSNR reductions of $0.20$ and $0.06$ dB. These results characterize the quality-computation trade-off observed when dynamics-conditioned internal revision is combined with learned continuation decisions.

\begin{figure*}[!ht]
    \centering
    \includegraphics[width=\textwidth]{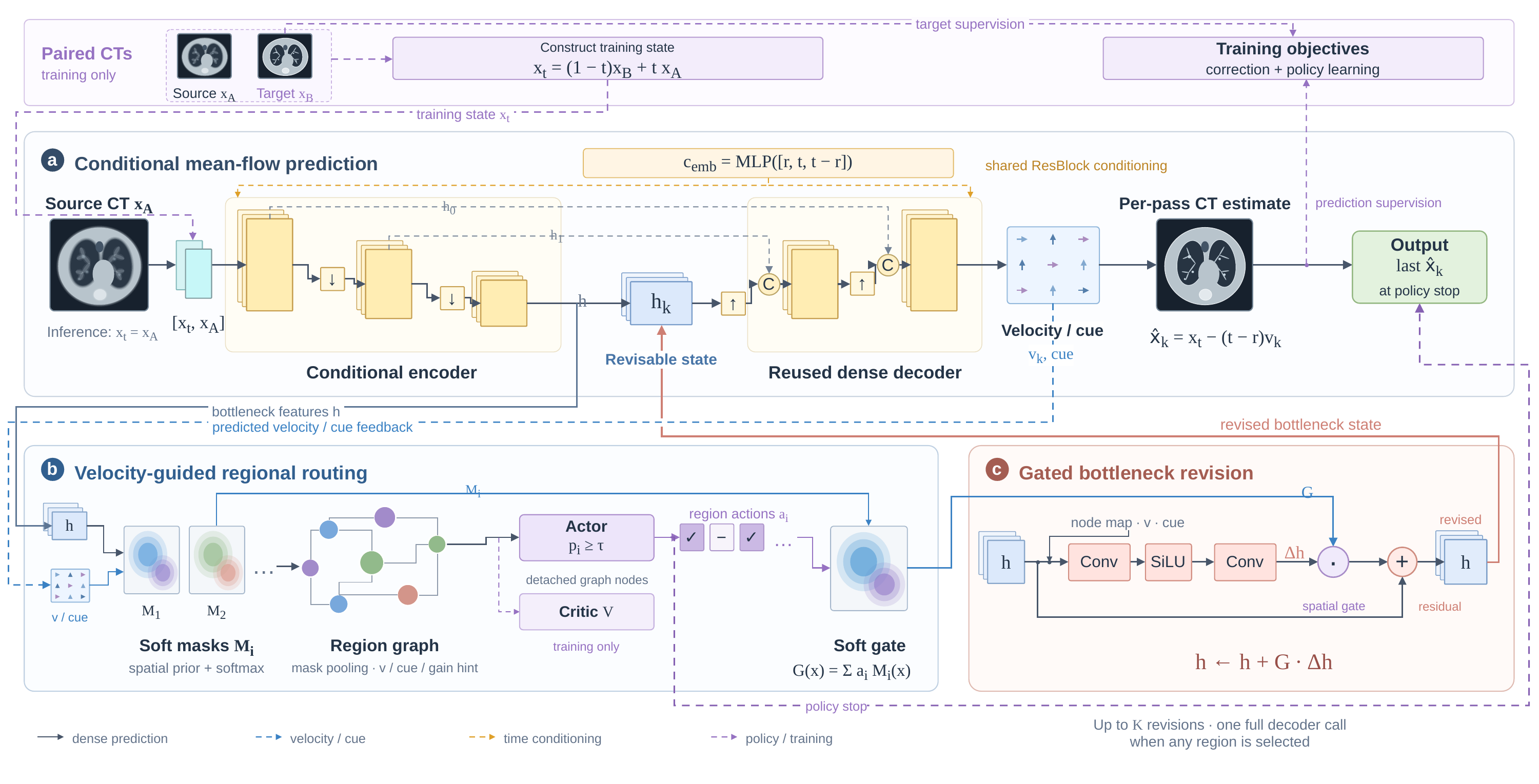}
    \caption{\textbf{IRCMF architecture overview.}
(a) A source-conditioned U-Net predicts signed interval
velocity and a learned change cue through a shared decoder.
(b) Predicted dynamics guide soft-region formation and graph
interaction, while a learned policy selects regions for
continuation.
(c) Selected regions gate recurrent bottleneck updates,
after which the shared decoder recomputes velocity.
Each executed revision uses a dense decoder evaluation.
The training-only paths provide supervision for the
prediction and continuation objectives.
Here $M_i$, $a_i$, and $G$ denote the router masks
$m_i^k$, actions $q_i^k$, and gate $g^k$ in the method.}
    \label{fig:ircmf_architecture}
\end{figure*}

\section{Related Work}
\label{sec:related_work}

\paragraph{Flow-based image enhancement and velocity modeling.}
Flow Matching~\cite{lipman2023flowmatching} represents generative transformations through learned instantaneous velocity fields, while MeanFlow~\cite{geng2025meanflow} introduces interval-average velocity prediction for one-step generation. Improved Mean Flows~\cite{geng2025improved} revisits its training formulation and conditioning. For image restoration, Reversing Flow~\cite{qin2025resflow} models deterministic degradation paths, PnP-Flow~\cite{martin2025pnpflow} incorporates flow priors into data-consistency updates, and Posterior-Mean Rectified Flow~\cite{ohayon2025pmrf} transports a posterior-mean estimate toward a high-quality image distribution. FlowTurbo~\cite{zhao2024flowturbo} refines velocity estimates to accelerate sampling, while RA-CMF~\cite{apurba2026racmf} learns regional refinement budgets for CT enhancement. These methods motivate velocity-based transformation and refinement, but differ in what is updated. Our work investigates how predicted signed velocity and a learned velocity-change cue can guide revisions of the internal features that generate the next velocity field.

\paragraph{Spatially adaptive features and regional interaction.}
Spatial variation in image degradation has motivated architectures that adapt feature processing to local content. SPAIR~\cite{purohit2021spair} combines distortion localization with spatially selective restoration, while MPRNet~\cite{zamir2021mprnet} exchanges information across progressive restoration stages. Degradation-aware feature perturbation~\cite{tian2025dfpir} further adapts internal representations according to degradation characteristics. Attention- and state-space-based restoration methods, including ACL~\cite{gu2025acl}, MaIR~\cite{li2025mair}, and MambaIRv2~\cite{guo2025mambairv2}, explore efficient feature interactions and local detail preservation. In CT enhancement, STAN-CT~\cite{selim2020stanct} and RadiomicGAN~\cite{selim2021radiomicgan} investigate image harmonization, while CLIP-guided denoising~\cite{cheng2024clipdenoising}, G2CR-FPN~\cite{zhou2024g2crfpn}, and ASCON~\cite{chen2023ascon} address noise reduction using pretrained representations, multiscale features, or anatomy-aware learning. Non-local networks~\cite{wang2018nonlocal} and graph attention networks~\cite{velickovic2018gat} further demonstrate how contextual information can be exchanged across spatial positions or learned nodes. These methods motivate feature adaptation and regional interaction without explicitly coupling them to the predicted transformation dynamics. In IRCMF, overlapping soft regions are informed by hidden image features and velocity-related signals, with graph communication contributing both to velocity prediction and recurrent bottleneck revision.

\paragraph{Iterative refinement and prediction-guided feedback.}
 Multi-stage restoration and recurrent inference use intermediate estimates to guide later predictions, but differ in the representation being updated. Restore, Assess, Repeat~\cite{chen2026rar} combines latent restoration with quality assessment and adaptive continuation. Region-Adaptive Sampling~\cite{liu2026ras} uses information from preceding diffusion steps to select regions for further updates, while TAP~\cite{zhu2026tap} selects token-level predictors using lightweight probes. These methods demonstrate that intermediate information can guide repeated computation. Our focus is feedback within a conditional velocity-producing network: predicted velocity and a learned velocity-change cue inform regional affinities, graph-guided gates revise bottleneck features, and the shared decoder produces a new velocity estimate. A learned actor-critic policy~\cite{mnih2016a3c} controls continuation through a training reward that balances full-image error reduction with decoder-call cost. Each executed revision still requires a dense decoder evaluation, so computational savings arise from skipping further evaluations rather than sparse decoding.

\section{Method}
\label{sec:method}

\subsection{Overview and Conditional Velocity Prediction}
\label{sec:method_overview}

% We propose \emph{velocity-guided internal revision}, a conditional enhancement framework in which a velocity predictor uses its own estimated transformation dynamics to revise the features that determine its next prediction. The key design choice is the \emph{location of feedback}. Rather than adjusting an image after the velocity has been decoded, we update the bottleneck representation of a shared U-Net~\cite{ronneberger2015unet} and recompute velocity using the same decoder. This creates a closed prediction-revision loop with three coupled operations: dynamics-conditioned velocity decoding, regional graph-guided feature revision, and learned continuation. The two graphs use distinct, normalized soft masks: decoder masks pool and return information used to contextualize the \emph{current} velocity, whereas revision masks form dynamic bottleneck nodes and gate the features used for the \emph{next} estimate. Neither mask represents a semantic segmentation or a hard crop. Figure~\ref{fig:ircmf_architecture} summarizes the prediction--revision loop; supplementary Fig.~\ref{supp:fig:ircmf_detailed} details the U-Net connections and both regional graphs.

We propose \emph{velocity-guided internal revision}, a conditional enhancement framework in which a velocity predictor uses its own estimated transformation dynamics to revise features for its next prediction. The key design choice is the \emph{location of feedback}. We update the bottleneck of a shared U-Net~\cite{ronneberger2015unet} and recompute velocity with the same decoder, rather than adjusting an image after decoding. This closed prediction-revision loop couples dynamics-conditioned velocity decoding, regional graph-guided feature revision, and learned continuation. The two graphs use distinct, normalized soft masks: decoder masks pool and return context for the \emph{current} velocity, while revision masks form dynamic bottleneck nodes and gate features for the \emph{next} estimate. Neither mask represents a semantic segmentation or a hard crop. Figure~\ref{fig:ircmf_architecture} summarizes the loop; supplementary Fig.~\ref{supp:fig:ircmf_detailed} details the U-Net connections and both regional graphs.

Let $x_A$ and $x_B$ denote paired source and target images and $d=x_A-x_B$ their signed displacement. For training, we define $x_t=x_B+td$, with $0\le r\le t\le1$. A conditional encoder computes bottleneck features $h^0$ and fixed skip features $s$. A shared decoder predicts signed interval velocity $v^k$ and a learned change cue $a^k$ at revision $k$:
\begin{equation}
\label{eq:cmf_state}
\begin{gathered}
(h^0,s)=\mathcal E_\theta(x_t,x_A,r,t),\\
(v^k,a^k)=\mathcal D_\theta(h^k,s,x_A,r,t),\\
\hat x_r^{\,k}=x_t-(t-r)v^k.
\end{gathered}
\end{equation}
The time embedding $c_{\mathrm{emb}}=\operatorname{MLP}([r,t,t-r])$ conditions every U-Net residual block and is shared across decoder evaluations of the same interval. At inference, $(x_t,r,t)=(x_A,0,1)$ in Eq.~\eqref{eq:cmf_state}. The interval-velocity parameterization objective uses direct paired supervision rather than the Jacobian-vector-product identity of MeanFlow~\cite{geng2025meanflow}. Source conditioning, encoder skips, and the decoding interval remain unchanged across revisions; only $h^k$ is revised.

\subsection{Dynamics-Conditioned Velocity Decoding}
\label{sec:method_decoder}

Each shared decoder evaluation produces features $f^k$, raw signed velocity $v_{\mathrm{raw}}^k$, projected features $F^k$, and learned change cue $a_g^k$. The decoder graph constructs $N_D=G_D^2$ region prototypes by spatially pooling $F^k$. For position $p$ and prototype $i$, its assignment logit combines normalized feature similarity, proximity to the prototype, and differences from prototype-averaged signed velocity and change cue:
\begin{equation}
\label{eq:cmf_decoder_masks}
\begin{aligned}
L_i^k(p)={}&\omega_{\mathrm{sim}}\operatorname{sim}(F_i^k,F^k(p))
-\rho_i(p)\\
&-\alpha_v(\delta_i^v(p))^2
-\alpha_a(\delta_i^a(p))^2,\\
M_i^k(p)={}&\frac{\mathbf1[i\in\mathcal N(p)]e^{L_i^k(p)}}
{\sum_{j\in\mathcal N(p)}e^{L_j^k(p)}}.
\end{aligned}
\end{equation}
Here $\omega_{\mathrm{sim}}$ is the feature-similarity scale, $\rho_i$ is a spatial-distance penalty, $\mathcal N(p)$ contains nearby prototypes, $\delta_i^v$ and $\delta_i^a$ are normalized differences from the prototype's dynamics, and the nonnegative $\alpha_v,\alpha_a$ are learned. The squared dynamics differences are clipped in the implementation. Thus $\sum_i M_i^k(p)=1$ on the lower-resolution working grid (Eq.~\eqref{eq:cmf_decoder_masks}). Using mask-normalized pooling $\langle u\rangle_{M_i}=\sum_pM_i(p)u(p)/\sum_pM_i(p)$, each decoder node embeds pooled $F^k$, source intensity, signed velocity, velocity magnitude, change cue and its magnitude, and region mass. Edges combine query-key similarity with relative signed velocity, change cue, and mask-weighted position, using learned regional affinities as in graph-attention methods~\cite{velickovic2018gat}:
\begin{equation}
\label{eq:cmf_decoder_edges}
\begin{gathered}
S_{ij}^{D,k}=\frac{(Q_Dz_{D,i}^k)^\top K_Dz_{D,j}^k}{\sqrt{d_D}}
+\phi_D(\Delta v_{ij},\Delta a_{ij},\Delta p_{ij}),\\
A_{ij}^{D,k}=\operatorname{softmax}_{j\in\mathcal N(i)}S_{ij}^{D,k}.
\end{gathered}
\end{equation}
In Eq.~\eqref{eq:cmf_decoder_edges}, $\mathcal N(i)$ is the immediate spatial neighborhood of node $i$, excluding $i$ itself. For the implemented local-difference graph, attention aggregates both signed node differences and their absolute magnitudes. Their learned edge messages are redistributed through $M^k$ to obtain $e_D^k$. Separately, a node-local transformation provides spatial context $c_D^k$ and a learned modulation score $b_D^k$; both are also scattered through the same masks. Hence the contextual correction $\Phi_D$ and edge projection $\Phi_E$ affect the current raw field through distinct local and relational paths:
\begin{equation}
\label{eq:cmf_decoder_velocity}
\begin{aligned}
v^k={}&v_{\mathrm{raw}}^k+
\left(\zeta_D+\sigma(b_D^k)\right)\odot\Phi_D(F^k,c_D^k,\\
&\qquad\quad v_{\mathrm{raw}}^k,a_g^k,x_A)
+\Phi_E(e_D^k)+\bar t\,a_d^k,\\
a^k={}&a_g^k+a_d^k.
\end{aligned}
\end{equation}
In Eq.~\eqref{eq:cmf_decoder_velocity}, the learned map $b_D^k$ modulates the contextual correction, $\zeta_D$ is its offset, and $\bar t=(r+t)/2$ is the interval midpoint multiplying the separate decoder cue $a_d^k$. Thus, this first graph changes the \emph{current} velocity estimate. The composite cue $a^k$ will inform internal revision; it is a learned predictor signal, not physical acceleration or measured error.

\subsection{Velocity-Guided Internal Revision}
\label{sec:method_revision}

The second graph is constructed afresh at every revision from the evolving bottleneck and the \emph{current} decoded dynamics. Let $\mathcal P$ denote adaptive average pooling to bottleneck resolution. Learned coarse heads supplement the decoded signals,
\begin{equation}
\label{eq:cmf_coarse}
\bar v^k=\mathcal P(v^k)+C_v(h^k),\qquad
\bar a^k=\mathcal P(a^k)+C_a(h^k).
\end{equation}
Using the coarse signals in Eq.~\eqref{eq:cmf_coarse}, the router predicts $N_R=J_RG_R^2$ \emph{overlapping} mask channels, with $J_R$ candidate memberships per spatial grid anchor. A pointwise projection of $[h^k,\mathcal P(x_A),\operatorname{std}(\bar v^k),\operatorname{std}(|\bar v^k|),\operatorname{std}(\bar a^k)]$ supplies learned logits $\ell_i^k$. A Gaussian-like spatial-anchor prior $\pi_i^{\mathrm{sp}}$ and a velocity-magnitude prior $\pi_i^{\mathrm{mag}}$ over the candidates regularize the assignments:
\begin{equation}
\label{eq:cmf_router_masks}
\begin{gathered}
m_i^k(p)=\operatorname{softmax}_i
\bigl(\ell_i^k(p)+\pi_i^{\mathrm{sp}}(p)
+\pi_i^{\mathrm{mag}}(p)\bigr),\\
\sum_i m_i^k(p)=1,\qquad
\operatorname{Pool}_i(u)=\frac{\sum_p m_i^k(p)u(p)}{\sum_p m_i^k(p)+\epsilon}.
\end{gathered}
\end{equation}
The masks in Eq.~\eqref{eq:cmf_router_masks} form a soft partition at each location, but permit a region to cover many positions and positions to contribute to multiple regions. They are neither anatomical labels nor independent crops. Mask-normalized pools of hidden features, source intensity, signed velocity, magnitude, and change cue, followed by a revision-stage embedding, define each router node $z_i^k$. Directed affinities compare both learned node features and regional transformation statistics:
\begin{equation}
\label{eq:cmf_router_affinity}
\begin{aligned}
S_{ij}^{R,k}={}&\frac{(Q_Rz_i^k)^\top K_Rz_j^k}{\sqrt{d_R}}
-\beta_v\left(\frac{\nu_i^k-\nu_j^k}{s_v}\right)^2\\
&-\beta_a\left(\frac{\eta_i^k-\eta_j^k}{s_a}\right)^2
+\widetilde b_j^k,\\
A_{ij}^{R,k}={}&\operatorname{softmax}_{j\ne i}S_{ij}^{R,k}.
\end{aligned}
\end{equation}
In Eq.~\eqref{eq:cmf_router_affinity}, $\nu_i^k$ and $\eta_i^k$ are pooled signed velocity and change cue, $s_v,s_a$ normalize nodewise variation, the nonnegative $\beta_v,\beta_a$ are learned, and $\widetilde b_j^k$ is a centered predicted donor gain. The graph exchanges pairwise difference messages and \emph{writes every updated node} back to the feature grid using the router masks:
\begin{equation}
\label{eq:cmf_router_messages}
\begin{gathered}
u_i^k=\sum_{j\ne i}A_{ij}^{R,k}
\Psi_R([z_i^k,z_j^k-z_i^k]),\\
\widetilde z_i^k=z_i^k+\Phi_R([z_i^k,u_i^k]),\\
c^k(p)=\sum_i m_i^k(p)W_R\widetilde z_i^k.
\end{gathered}
\end{equation}
For binary node actions $q_i^k$ from the regional actor, the \emph{same masks} produce the spatial continuation gate $g^k(p)=\sum_iq_i^km_i^k(p)$. The contextual map $c^k$ in Eq.~\eqref{eq:cmf_router_messages} incorporates all router nodes; actions gate the subsequent feature increment, not graph-message computation. A shared recurrent cell revises bottleneck features while the source, skip features, and decoder parameters remain fixed:
\begin{equation}
\label{eq:cmf_revision}
\begin{gathered}
h^{k+1}(p)=h^k(p)+\gamma g^k(p)\,
\tanh R_\theta(\xi^k(p)),\\
\xi^k=[h^k,c^k,\bar v^k,\bar a^k],\\
(v^{k+1},a^{k+1})=\mathcal D_\theta(h^{k+1},s,x_A,r,t).
\end{gathered}
\end{equation}
In Eq.~\eqref{eq:cmf_revision}, the update scale $\gamma$ controls the magnitude of the feature increment. The revision therefore acts \emph{before} velocity recomputation. Locally, $v^{k+1}-v^k\approx J_{\mathcal D^v,h}(h^{k+1}-h^k)$ explains how revised features alter the next field without assuming an improvement at every step. Since $\sum_i m_i^k(p)=1$, choosing all nodes gives $g^k(p)=1$; regional selectivity arises when the policy chooses only a subset. Thus, all-selected image losses train mask-dependent graph context and the revision cell, but do not directly supervise selective gating. The decoder nevertheless runs densely whenever any node acts. Unlike further flow integration, $(x_t,r,t)$ remains fixed while both masks and velocity are recomputed after each active revision.

\subsection{Learning Revision and Continuation}
\label{sec:method_learning}

Paired training supervises initial, intermediate, and final endpoint predictions along the all-selected revision trajectory, along with velocity from one decoder evaluation on randomly sampled finite intervals. Write $E_{\mathrm{all}}^k=\operatorname{MSE}(\hat x_r^{\,k},x_B)$ for this source-to-target trajectory at $(r,t)=(0,1)$, and let $d=x_A-x_B$. The principal losses are
\begin{equation}
\label{eq:cmf_losses}
\begin{gathered}
\mathcal L_{\mathrm{end}}=E_{\mathrm{all}}^K+\lambda_0E_{\mathrm{all}}^0
+\lambda_m K^{-1}\!\sum_{k=1}^{K}E_{\mathrm{all}}^k,\\
\mathcal L_{\mathrm{int}}=\mathbb E_{r,t}
\|v_\theta(x_t,x_A,r,t)-d\|_2^2.
\end{gathered}
\end{equation}
Regional error increases are penalized using the \emph{router masks} to pool squared error before and after each all-selected revision. In each comparison below, $E_i^k$ and $E_i^{k+1}$ use the same step-$k$ mask $m_i^k$. During this auxiliary loss, mask weights and the earlier error are detached, so the penalty backpropagates only through the updated prediction. A node gain head is trained on the clipped observed decrease in the same mask-pooled error. The coarse velocity and change cue are additionally supervised against pooled displacement and the detached difference between successive velocity estimates:
\begin{equation}
\label{eq:cmf_aux}
\begin{aligned}
\mathcal L_{\mathrm{reg}}&=\mathbb E_{k,i}
[\,E_i^{k+1}-\operatorname{sg}(E_i^k)\,]_+,\\
\mathcal L_{\mathrm{gain}}&=\mathbb E_{k,i}
\operatorname{Huber}(b_i^k,\operatorname{sg}(\hat g_i^k)),\\
\mathcal L_{\mathrm{coarse}}&=\|\bar v^0+\chi_a\bar a^0-\mathcal P(d)\|_2^2,\\
\mathcal L_{\mathrm{change}}&=\operatorname{Huber}(\bar a^0,
\operatorname{sg}(\mathcal P(v^1-v^0))).
\end{aligned}
\end{equation}
Here $\hat g_i^k=\operatorname{clip}(\kappa(E_i^k-E_i^{k+1}),-c,c)$ is the scaled regional error decrease, $\chi_a$ weights the coarse change cue, and $\operatorname{sg}$ stops gradients. These supervision signals are available only during training. The complete correction objective is a weighted sum of Eqs.~\eqref{eq:cmf_losses} and~\eqref{eq:cmf_aux}; supplementary Tab.~\ref{supp:tab:training} specifies its weights and training schedule.

A regional actor predicts Bernoulli continuation probabilities $\pi_i^k$ and a critic predicts $V_i^k$, following actor-critic learning~\cite{konda1999actorcritic}. During policy training, actions are sampled from the current policy. Write $E_{\mathrm{pol}}^k$ for the full-image error along this sampled trajectory; a stage reward trades its error reduction against the cost of any additional \emph{dense} decoder call:
\begin{equation}
\label{eq:cmf_policy_reward}
\begin{aligned}
R_k={}&\kappa(E_{\mathrm{pol}}^k-E_{\mathrm{pol}}^{k+1})
-\lambda_w\mathbf1[\textstyle\sum_i q_i^k>0],\\
G_k={}&\sum_{j=k}^{K-1}R_j,\\
\widehat A_i^k={}&\operatorname{clip}_{[-c_{\mathrm{adv}},c_{\mathrm{adv}}]}(G_k-\operatorname{sg}(V_i^k)),\\
\mathcal L_{\mathrm{AC}}={}&
-\mathbb E[\log p_\theta(q_i^k)\widehat A_i^k]\\
&+\lambda_V\mathbb E[\operatorname{Huber}(V_i^k,G_k)]
-\lambda_H\mathcal H(\pi).
\end{aligned}
\end{equation}
In Eq.~\eqref{eq:cmf_policy_reward}, all eligible regions receive the same image-level return, not individually estimated causal rewards; $c_{\mathrm{adv}}$ bounds the policy advantage. The policy uses detached router node embeddings, and its score-function objective is on-policy actor-critic, not PPO. Total training combines the correction and policy objectives. In the reported run, the sampled trajectory has no separate image loss; its target-derived reward is detached from the feature and revision networks. At inference, actions are thresholded at $\tau$; a stopped node cannot restart, and if no node acts the procedure terminates. The policy controls the \emph{number} of complete decoder evaluations, not the spatial cost within one evaluation. Architecture, graph, and policy settings are listed in supplementary Tables~\ref{supp:tab:architecture}, \ref{supp:tab:graph}, and~\ref{supp:tab:training}. Algorithm~\ref{alg:cmf_inference} summarizes target-free inference, with $v^{\mathrm{last}}$ denoting the latest decoded velocity.

\begin{algorithm}[t]
\caption{Velocity-guided internal revision}
\label{alg:cmf_inference}
\begin{algorithmic}[1]
\Require Source $x_A$, budget $K$
\State $(x_t,r,t)\gets(x_A,0,1)$
\State $(h^0,s)\gets\mathcal E_\theta(x_t,x_A,r,t)$
\State $(v^0,a^0)\gets\operatorname{DecodeGraph}(h^0,s,x_A,r,t)$
\State $\mathcal A\gets\{1,\ldots,N_R\}$
\For{$k=0,\ldots,K-1$}
  \State $(\bar v^k,\bar a^k)\gets\operatorname{Coarse}(v^k,a^k,h^k)$
  \State $(m^k,c^k,\pi^k)\gets\operatorname{RouterMasksGraphActor}(h^k,x_A,\bar v^k,\bar a^k,k)$
  \State $q_i^k\gets\mathbf1[i\in\mathcal A]\mathbf1[\pi_i^k\ge\tau]$
  \If{$\sum_iq_i^k=0$} \State \textbf{break} \EndIf
  \State $g^k(p)\gets\sum_iq_i^k m_i^k(p)$
  \State $h^{k+1}\gets\operatorname{Revise}(h^k,c^k,\bar v^k,\bar a^k,g^k)$
  \State $(v^{k+1},a^{k+1})\gets\operatorname{DecodeGraph}(h^{k+1},s,x_A,r,t)$
  \State $\mathcal A\gets\{i:q_i^k=1\}$
\EndFor
\State \Return $x_t-(t-r)v^{\mathrm{last}}$
\end{algorithmic}
\end{algorithm}

\section{Experiments}
\label{sec:experiments}
We evaluate velocity-guided internal revision on paired-image subsets of the National Lung Screening Trial (NLST) and Low Dose CT Image and Projection Data (LDCT-and-Projection-data), examining enhancement accuracy, architectural components, and the quality-computation trade-off of regional continuation.

\subsection{Experimental Setup}
\label{sec:exp_setup}

\paragraph{National Lung Screening Trial (NLST) Dataset.} We curate paired axial chest CT slices from NLST~\cite{nlst2013}. For selected GE examinations, \texttt{STANDARD}-kernel images serve as the source and anatomically corresponding \texttt{LUNG}-kernel images as the target. Each pair represents the same scan, with kernel-dependent differences in detail, noise, and texture (Tab.~\ref{tab:data_split}).

\paragraph{Low Dose CT Image and Projection Dataset.} We curate 49 Siemens chest CT series from LDCT-and-Projection-data~\cite{mccollough2020ldct}: 18 SOMATOM Definition Flash and 31 SOMATOM Definition AS+ series reconstructed with the \texttt{B50f} kernel. Each simulated low-dose image is paired with its routine-dose reference, differing in dose-related noise rather than reconstruction kernel (Tab.~\ref{tab:data_split}).

\paragraph{NLSTseg tumor-ROI subset.}
For region-specific evaluation, we use five patients and 71 tumor-positive axial slices from NLSTseg~\cite{chen2025nlstseg}, which provides pixel-level lesion annotations derived from NLST. The masks define tumor regions for fidelity measurements and radiomic extraction, separately from the full-image test cohorts. Supplementary Tab.~\ref{supp:tab:subjects} lists the subject identifiers. 

\begin{table}[t]
\centering
\caption{Patient-disjoint subsets of NLST and LDCT-and-Projection-data. Entries give patients / paired axial slices.}
\label{tab:data_split}
\setlength{\tabcolsep}{1pt}
\small
\begin{tabular*}{\columnwidth}{@{\extracolsep{\fill}}lccc@{}}
\toprule
Dataset & \shortstack{Training\\(Patient/\\Slices)}
        & \shortstack{Validation\\(Patient/\\Slices)}
        & \shortstack{Test\\(Patient/\\Slices)} \\
\midrule
\shortstack[l]{NLST\\(GE)} & 29 / 15,095 & 7 / 3,579 & 3 / 1,695 \\
\midrule
\shortstack[l]{LDCT\\(Siemens)} & 39 / 13,669 & 5 / 1,642 & 5 / 1,699 \\
\bottomrule
\end{tabular*}
\end{table}

\begin{table*}[th!]
\centering
\caption{Full-image enhancement comparison on NLST and LDCT-and-Projection-data (LDCT).
PSNR and SSIM are slice-level mean $\pm$ standard deviation.}
\label{tab:enhancement}
\setlength{\tabcolsep}{5pt}
\small
\begin{threeparttable}
\begin{tabular*}{\textwidth}{@{\extracolsep{\fill}}llccrr@{}}
\toprule
Dataset & Method & PSNR (dB) $\uparrow$ & SSIM $\uparrow$ & NFE $\downarrow$ & Latency (ms) $\downarrow$ \\
\midrule
\multirow{12}{*}{NLST} & Input & $22.29\pm2.36$ & $0.65\pm0.09$ & -- & -- \\
 & CNN-baseline~\cite{lee2019ctconversion} & $24.65\pm2.90$ & $0.77\pm0.11$ & \textbf{1} & \textbf{2.66} \\
 & STAN-CT~\cite{selim2020stanct} & $29.25\pm4.03$ & $0.87\pm0.14$ & \textbf{1} & \underline{3.88} \\
 & RadiomicGAN~\cite{selim2021radiomicgan} & $28.29\pm3.61$ & $0.86\pm0.14$ & \textbf{1} & 16.41 \\
 & CMF~\cite{apurba2026racmf} & $31.41\pm2.13$ & $0.91\pm0.02$ & \textbf{1} & 14.41 \\
 & ACL~\cite{gu2025acl} & $33.59\pm2.83$ & $0.95\pm0.01$ & \textbf{1} & 373.23 \\
 & MaIR-S~\cite{li2025mair} & $32.53\pm2.51$ & $0.95\pm0.01$ & \textbf{1} & 620.62 \\
 & MambaIRv2~\cite{guo2025mambairv2} & $33.67\pm2.78$ & $0.95\pm0.01$ & \textbf{1} & 6580.46 \\
 & RA-CMF~\cite{apurba2026racmf} & $33.74\pm2.75$ & $0.95\pm0.02$ & 16 & 463.62 \\
\addlinespace[1pt]
\cmidrule(lr){2-6}
 & IRCMF (OP)* & $35.84\pm3.36$ & $\underline{0.96\pm0.01}$ & \textbf{1} & 15.61 \\
 & IRCMF (LP)* & $\underline{36.04\pm3.41}$ & $\underline{0.96\pm0.01}$ & \underline{4.42} & 51.82 \\
 & IRCMF (AR)* & $\mathbf{36.24\pm3.49}$ & $\mathbf{0.97\pm0.01}$ & 9 & 97.00 \\
\midrule
\multirow{10}{*}{LDCT} & Input & $22.22\pm2.25$ & $0.59\pm0.09$ & -- & -- \\
 & CLIPDenoising~\cite{cheng2024clipdenoising} & $22.87\pm2.41$ & $0.68\pm0.08$ & \textbf{1} & \textbf{8.97} \\
 & G2CR-FPN~\cite{zhou2024g2crfpn} & $28.07\pm1.54$ & $0.77\pm0.05$ & \textbf{1} & 741.96 \\
 & CMF~\cite{apurba2026racmf} & $29.60\pm1.76$ & $0.78\pm0.06$ & \textbf{1} & \underline{14.04} \\
 & ASCON~\cite{chen2023ascon} & $30.77\pm1.41$ & $\underline{0.83\pm0.05}$ & \textbf{1} & 23.53 \\
 & RA-CMF~\cite{apurba2026racmf} & $30.85\pm1.81$ & $0.82\pm0.05$ & 16 & 260.15 \\
 & MaIR-S~\cite{li2025mair} & $31.17\pm1.71$ & $0.82\pm0.05$ & \textbf{1} & 1154.63 \\
\addlinespace[1pt]
\cmidrule(lr){2-6}
 & IRCMF (OP)* & $32.37\pm1.92$ & $\mathbf{0.84\pm0.05}$ & \textbf{1} & 15.52 \\
 & IRCMF (LP)* & $\underline{32.39\pm1.93}$ & $\mathbf{0.84\pm0.05}$ & \underline{3.45} & 41.95 \\
 & IRCMF (AR)* & $\mathbf{32.45\pm1.97}$ & $\mathbf{0.84\pm0.05}$ & 9 & 97.12 \\
\bottomrule
\end{tabular*}
\begin{tablenotes}[flushleft]
\footnotesize
\item \textit{Note.} OP = one pass; AR = all revisions; LP = learned policy; * marks our method. Bold and underline mark the best and second-best distinct displayed means per dataset and metric; ties share markings (higher PSNR/SSIM, lower NFE/latency). NFE counts baseline model evaluations or IRCMF dense decoder calls; Input is excluded from NFE ranking.
\end{tablenotes}
\end{threeparttable}
\end{table*}

% \paragraph{Implementation details.}
% We train separate models from scratch on both datasets with the same conditional velocity U-Net, decoder-side soft-region graph, bottleneck router graph, and shared recurrent reviser. We use 64 decoder graph regions, 128 overlapping router regions, and up to eight additional decoder evaluations. Optimization uses AdamW~\cite{loshchilov2019adamw} with learning rates of $10^{-4}$ for the enhancement model and $3\times10^{-4}$ for the actor-critic heads, an effective batch size of 32, and bfloat16 precision on two GPUs. Geometrically aligned slices are processed at $512\times512$ resolution. Checkpoints are selected by validation patient-macro PSNR in all-revisions mode (epoch 29 for NLST; epoch 30 for LDCT).

\paragraph{Implementation details.}
We train separate IRCMF models from scratch on both datasets using the same architecture and AdamW optimizer~\cite{loshchilov2019adamw}. Checkpoints are selected by validation patient-macro PSNR under all revisions. Detailed architecture, optimization, preprocessing, and evaluation settings are provided in Sec.~\ref{supp:sec:implementation} of the supplementary material.

\paragraph{Comparison methods.}
We compare with a CNN baseline~\cite{lee2019ctconversion}, STAN-CT~\cite{selim2020stanct}, RadiomicGAN~\cite{selim2021radiomicgan}, CMF~\cite{apurba2026racmf}, RA-CMF~\cite{apurba2026racmf}, ACL~\cite{gu2025acl}, MaIR-S~\cite{li2025mair}, MambaIRv2~\cite{guo2025mambairv2}, CLIPDenoising~\cite{cheng2024clipdenoising}, G2CR-FPN~\cite{zhou2024g2crfpn}, and ASCON~\cite{chen2023ascon}. Available baselines differ between datasets (Tab.~\ref{tab:enhancement}); supplementary Tables~\ref{supp:tab:common-protocol} and~\ref{supp:tab:coverage} document the common NLST protocol and training provenance.

\paragraph{Metrics and comparison protocol.}
We evaluate peak signal-to-noise ratio (PSNR, dB) and structural similarity (SSIM)~\cite{wang2004ssim} against paired references, reporting slice-level mean $\pm$ sample standard deviation unless noted. Computational cost includes synchronized batch-one CUDA latency and NFE. For IRCMF, NFE counts dense decoder evaluations including the initial prediction; one baseline model evaluation need not have equivalent cost. Supplementary Tab.~\ref{supp:tab:parameters} reports parameter counts.

\subsection{Quantitative Comparison}
\label{sec:exp_results}
Table~\ref{tab:enhancement} compares the baselines with one pass, all eight revisions, and learned-policy inference of the \emph{same trained IRCMF model}. On NLST, all revisions achieve $36.24$ dB PSNR and $0.97$ SSIM versus $35.84$ dB and $0.96$ in one pass; on LDCT, the corresponding values are $32.45$ versus $32.37$ dB and $0.84$ versus $0.84$ SSIM. The strongest non-IRCMF PSNR is $33.74$ dB for RA-CMF on NLST and $31.17$ dB for MaIR-S on LDCT, showing different gains across enhancement settings.

The learned policy retains $36.04$ dB / $0.96$ SSIM on NLST at 4.42 decoder evaluations and $32.39$ dB / $0.84$ SSIM on LDCT at 3.45, reducing PSNR by only $0.20$ and $0.06$ dB relative to all revisions. Patient-macro PSNR for all revisions and the policy is $36.24$ and $36.04$ dB on NLST, and $32.46$ and $32.40$ dB on LDCT. With only three and five independent test patients, slice-level deviations are not population-level confidence intervals.

\subsection{Tumor ROI and Radiomic Feature Evaluation}
\label{sec:exp_tumor_roi}

We evaluate tumor-region fidelity and radiomic preservation on five NLSTseg patients (71 tumor-positive slices). Tumor-only PSNR and SSIM measure agreement within the annotations. Following established definitions~\cite{zwanenburg2020ibsi}, PyRadiomics~\cite{vangriethuysen2017pyradiomics} extracts first-order statistics and five texture feature classes -- GLCM~\cite{haralick1973textural}, GLRLM~\cite{galloway1975texture}, GLSZM~\cite{thibault2009texture}, GLDM~\cite{sun1983neighboring}, and NGTDM~\cite{amadasun1989textural}. Concordance correlation coefficient (CCC)~\cite{lin1989ccc} measures feature agreement with paired targets.

As shown in Tab.~\ref{tab:roi_fidelity}, all-revisions inference achieves the highest tumor-ROI PSNR and SSIM, whereas one-pass inference obtains the highest overall radiomic CCC. Detailed class-wise results are provided in Tab.~\ref{supp:tab:radiomics} of the supplementary material.

\begin{table}[!htbp]
\centering
\caption{Overall radiomic CCC and tumor-ROI fidelity on slices from the five-patient NLSTseg subset of NLST. Entries are mean $\pm$ standard deviation.}
\label{tab:roi_fidelity}

\normalsize
\setlength{\tabcolsep}{0.2pt}
\renewcommand{\arraystretch}{1.12}
\begin{threeparttable}
\begin{tabularx}{\columnwidth}{@{}
    >{\raggedright\arraybackslash}p{0.26\columnwidth}
    >{\centering\arraybackslash}p{0.24\columnwidth}
    >{\centering\arraybackslash}p{0.24\columnwidth}
    >{\centering\arraybackslash}X
@{}}
\toprule
Method & \shortstack{Overall CCC\\$\uparrow$}
       & \shortstack{ROI PSNR\\(dB) $\uparrow$}
       & \shortstack{ROI SSIM\\$\uparrow$} \\
\midrule

Input & $0.51\pm0.24$ & $17.81\pm1.84$ & $0.53\pm0.13$ \\
CNN-baseline & $0.71\pm0.22$ & $20.29\pm1.89$ & $0.69\pm0.10$ \\
STAN-CT & $0.75\pm0.25$ & $26.58\pm2.48$ & $0.88\pm0.16$ \\
RadiomicGAN & $0.92\pm0.08$ & $27.30\pm2.79$ & $0.91\pm0.07$ \\
CMF & $0.89\pm0.15$ & $29.57\pm1.98$ & $0.94\pm0.03$ \\
RA-CMF & $0.93\pm0.09$ & $31.94\pm2.64$ & $0.97\pm0.03$ \\
ACL & $0.90\pm0.15$ & $33.19\pm2.50$ & $0.97\pm0.02$ \\
MaIR-S & \underline{$0.94\pm0.06$} & $32.04\pm2.26$ & $0.96\pm0.03$ \\
MambaIRv2 & $0.91\pm0.15$ & $32.50\pm2.66$ & $0.95\pm0.04$ \\
\midrule
IRCMF (OP)* & $\boldsymbol{0.95\pm0.07}$ & $34.43\pm2.51$ & \underline{$0.98\pm0.02$} \\
IRCMF (AR)* & $0.93\pm0.11$ & $\boldsymbol{37.07\pm2.82}$ & $\boldsymbol{0.99\pm0.01}$ \\
IRCMF (LP)* & $0.93\pm0.12$ & \underline{$35.92\pm2.55$} & \underline{$0.98\pm0.02$} \\
\bottomrule
\end{tabularx}

\begin{tablenotes}[flushleft]
\footnotesize
\item \textit{Note.} OP = one pass; AR = all revisions; LP = learned policy; * marks our method. Bold and underline indicate the highest and second-highest mean for each metric; ties receive the same marking.
\end{tablenotes}
\end{threeparttable}
\end{table}

% Requires \usepackage{placeins} in the main manuscript preamble.
% \FloatBarrier

\subsection{Ablation Studies}
\label{sec:exp_ablation}
Table~\ref{tab:exp_ablation} builds IRCMF progressively from \emph{our conditional CMF backbone}, not an external baseline. Each row retains the preceding components; all five configurations are trained independently for 10 epochs and tested on identical NLST slices. Velocity-guided soft-region masks raise PSNR from $31.20$ to $31.81$ dB but lower SSIM from $0.92745$ to $0.89619$. The learned velocity-change cue then raises performance to $33.52$ dB and $0.95062$ SSIM, followed by regional graph communication at $34.41$ dB and $0.95429$. Recurrent internal revision reaches $35.03$ dB and $0.95744$, gains of $3.83$ dB and $0.0300$ SSIM over our backbone. This 10-epoch study is distinct from the full training runs in Tab.~\ref{tab:enhancement}. Supplementary Tab.~\ref{supp:tab:cue_ablation} reports the matched cue-removal ablation with all revisions enabled.

\begin{table}[t]
\centering
\caption{Cumulative component ablation on NLST. Every configuration, including the complete IRCMF model, is independently trained for 10 epochs using the same NLST data split and evaluated on the same test slices.}
\label{tab:exp_ablation}
\setlength{\tabcolsep}{3pt}
\normalsize
\begin{tabular*}{\columnwidth}{@{\extracolsep{\fill}}>{\raggedright\arraybackslash}p{0.47\columnwidth}cc@{}}
\toprule
Components & PSNR (dB) $\uparrow$ & SSIM $\uparrow$ \\
\midrule
CMF backbone & 31.1957 & 0.92745 \\
$+$ Velocity-guided soft-region masks & 31.8131 & 0.89619 \\
$+$ Learned velocity-change cue & 33.5170 & 0.95062 \\
$+$ Regional graph communication & 34.4142 & 0.95429 \\
\midrule
$+$ Internal revisions (full IRCMF) & \textbf{35.0262} & \textbf{0.95744} \\
\bottomrule
\end{tabular*}
\end{table}

\subsection{Adaptive Revision Analysis}
\label{sec:exp_adaptive}
The policy selects continue/stop actions for overlapping router regions. Selected masks gate bottleneck updates before dense velocity recomputation. Supplementary Fig.~\ref{supp:fig:spatial-continuation} and Tab.~\ref{supp:tab:gate-analysis} show spatial gates and full-test continuation behavior. Computation is reduced by skipping evaluations rather than sparsifying decoder calls.

Relative to nine all-revisions evaluations, the policy averages 4.42 on NLST and 3.45 on LDCT, reducing latency from 97.00 to 51.82 ms and from 97.12 to 41.95 ms (Tab.~\ref{tab:enhancement}). Mean cumulative node activations are 251.15 and 116.61 per slice, distinct from unique regions or decoder evaluations.

\subsection{Qualitative Evaluation}
\label{sec:exp_qual}

Figure~\ref{fig:tumor_roi_examples} illustrates tumor-region enhancement on three NLSTseg cases. Compared with the paired targets, the input images exhibit smoother local textures and less pronounced structural details. IRCMF produces outputs that more closely resemble the target appearance, particularly around tumor boundaries and adjacent pulmonary structures. The green circles highlight tumor-containing regions for visual comparison rather than representing exact segmentation boundaries. The displayed PSNR and SSIM values measure fidelity within the annotated tumor regions, while Tab.~\ref{tab:roi_fidelity} reports aggregate results. Additional qualitative comparisons across NLST reconstruction-kernel enhancement, LDCT dose enhancement, and NLSTseg tumor-region enhancement are presented in Sec.~\ref{supp:sec:qualitative} of the supplementary material.

\section{Discussion and Limitations}
\label{sec:discussion}

Our results demonstrate that predicted transformation dynamics can guide internal feature revision to improve conditional image enhancement. Unlike iterative image-space correction, IRCMF revises the representations responsible for subsequent velocity predictions while retaining the same conditional state and shared decoder. The benefit of additional revisions differs across enhancement tasks, suggesting that a fixed revision schedule is not equally necessary for every input. The matched change-cue ablation in supplementary Tab.~\ref{supp:tab:cue_ablation} supports the contribution of learned dynamics conditioning, while the spatial-gate analysis in supplementary Sec.~\ref{supp:sec:spatial-gate} demonstrates how the continuation policy progressively reduces regional activity.

The learned policy retains most of the full-revision enhancement quality with substantially fewer decoder evaluations and lower inference latency. This demonstrates the practical value of adapting revision depth rather than applying the complete schedule uniformly. Although regional gates control where internal features are updated, each executed decoder evaluation remains dense, with computational savings arising from early termination. The current evaluation covers two enhancement settings with limited independent test patients; broader cohorts and acquisition conditions will further establish the generalizability of the approach.

\section{Conclusion}
\label{sec:conclusion}

We introduced IRCMF, a conditional image-enhancement framework that uses signed velocity and learned change cues to guide soft-region interactions and recurrent internal feature revision. A shared decoder recomputes velocity after each revision, while a learned policy controls continuation. Experiments on curated NLST and LDCT-and-Projection-data subsets show improved enhancement fidelity, with learned continuation retaining most of the full-revision quality at fewer dense decoder evaluations and lower latency. The matched change-cue ablation supports the cue's contribution, while full-test spatial-gate measurements characterize how the policy distributes and terminates regional updates. Evaluation on five NLSTseg patients additionally assesses tumor-region fidelity and radiomic agreement. Together, these findings support dynamics-conditioned internal feedback for conditional enhancement with adaptive computation. Broader validation across patient cohorts and acquisition settings remains an important direction for future work.

\section*{LLM Usage Statement}
The authors used generative AI tools, including ChatGPT and Codex, to assist with language editing, literature discovery, mathematical exposition, and source-code preparation. The authors independently verify the mathematical formulations, citations, experimental records, and final claims and retain full responsibility for the accuracy and integrity of the manuscript.

{\small
\bibliographystyle{ieeenat_fullname}
\bibliography{references}}

% ============================
% Supplementary Material
% ============================
\clearpage
\onecolumn
\input{supplementary}

\end{document}

%% file: supplementary.tex
% \documentclass[10pt,twocolumn,letterpaper]{article}
% \usepackage[]{cvpr}
% \usepackage{mathtools,tabularx,array,microtype,needspace}
% \usepackage[section]{placeins}
% \usepackage{float}
% \usepackage{placeins}
% \captionsetup{hypcap=false}
% \captionsetup[table]{font=normalsize,justification=raggedright,singlelinecheck=false}
% \definecolor{cvprblue}{rgb}{0.21,0.49,0.74}
% \usepackage[pagebackref,breaklinks,colorlinks,allcolors=cvprblue]{hyperref}
% \usepackage[percent]{overpic}
% \usepackage{subcaption}
% \def\paperID{*****}\def\confName{CVPR}\def\confYear{2027}
% \newcolumntype{Y}{>{\raggedright\arraybackslash}X}
% \renewcommand{\thesection}{S\arabic{section}}
% \renewcommand{\thesubsection}{\thesection.\arabic{subsection}}
% \renewcommand{\thetable}{S\arabic{table}}
% \renewcommand{\thefigure}{S\arabic{figure}}
% \renewcommand{\theequation}{S\arabic{equation}}
% \hyphenation{IRCMF PSNR SSIM NLST GLCM NGTDM}
% \title{Supplementary Material}
% % \author{}
% % \date{}
% \begin{document}
% \raggedbottom
% \onecolumn
% \maketitle

% ============================
% Supplementary Material
% ============================

\setcounter{section}{0}
\setcounter{subsection}{0}
\setcounter{table}{0}
\setcounter{figure}{0}
\setcounter{equation}{0}

\renewcommand{\thesection}{S\arabic{section}}
\renewcommand{\thesubsection}{\thesection.\arabic{subsection}}
\renewcommand{\thetable}{S\arabic{table}}
\renewcommand{\thefigure}{S\arabic{figure}}
\renewcommand{\theequation}{S\arabic{equation}}

\raggedbottom

\begin{center}
    {\LARGE\bfseries Supplementary Material}
\end{center}

\vspace{1em}

This document accompanies the main manuscript and contains additional analyses of regional selection and velocity-related cues, followed by the architectural, optimization, preprocessing, and evaluation details required to reproduce the reported setup. It also lists subject-level splits and parameter counts. Full-image enhancement, tumor-ROI fidelity, and the principal component ablation are presented in the main paper and are not repeated here.

\section{Additional Regional Analyses}
\label{supp:sec:regional}

This section examines the contribution of the learned
change cue and the spatial continuation behavior of
IRCMF. The change-cue ablation uses matched training
configurations, while the spatial-gate analysis uses
the validation-selected full-model checkpoints on
the complete NLST and LDCT test sets.

\subsection{Contribution of Dynamics Conditioning and Regional Interaction}
\label{supp:sec:cue-ablation}

The learned change cue and dynamics-conditioned regional
interactions provide additional conditioning signals for
velocity estimation and internal revision. To evaluate their
contributions, we compare IRCMF with configurations that
remove or replace individual components. All configurations
follow the same training and evaluation protocol, with all
internal revisions enabled and nine decoder evaluations.
This comparison uses the 10-epoch NLST ablation setting,
distinct from the validation-selected full-model experiments.

As shown in Tab.~\ref{supp:tab:cue_ablation}, IRCMF
achieves $35.0262$ dB PSNR and $0.95744$ SSIM,
compared with $34.1792$ dB and $0.95071$ SSIM
without the learned change-cue pathway. Including the cue
improves PSNR by $0.8470$ dB and SSIM by $0.00673$.
Uniform full-image revision reduces PSNR by $0.8558$ dB,
while removing signed-velocity routing, speed routing,
or router graph messages also decreases enhancement
fidelity. Replacing learned regional masks with fixed
masks and local nodes reduces PSNR by $0.6408$ dB.
These results demonstrate the contribution of dynamics
conditioning and regional interaction components under
matched training conditions.

\begin{table}[!htbp]
\raggedright
\setlength{\tabcolsep}{4pt}
\renewcommand{\arraystretch}{1.12}
\caption{Dynamics conditioning and regional interaction
ablations on NLST. All configurations use the same
10-epoch training and evaluation protocol with
nine decoder evaluations. Changes are relative to
the complete regional model.}
\label{supp:tab:cue_ablation}
\begin{tabularx}{\linewidth}{@{}Yrrrr@{}}
\toprule
Configuration & PSNR (dB) $\uparrow$ & SSIM $\uparrow$
& $\Delta$PSNR & $\Delta$SSIM\\
\midrule
IRCMF (All revisions)
& \textbf{35.0262} & \textbf{0.95744} & -- & --\\
Uniform full-image revision
& 34.1704 & 0.95228 & $-0.8558$ & $-0.00516$\\
Without learned change-cue pathway
& 34.1792 & 0.95071 & $-0.8470$ & $-0.00673$\\
Without signed-velocity routing
& 34.3716 & 0.95294 & $-0.6546$ & $-0.00450$\\
Without speed routing
& 34.2790 & 0.95256 & $-0.7472$ & $-0.00488$\\
Without router graph messages
& 34.3499 & 0.95253 & $-0.6763$ & $-0.00491$\\
Fixed masks with local nodes
& 34.3854 & 0.95405 & $-0.6408$ & $-0.00339$\\
\bottomrule
\end{tabularx}
\end{table}

\subsection{Spatial-Gate Analysis}
\label{supp:sec:spatial-gate}

\begin{figure}[t]
\centering

\begin{subfigure}[t]{0.49\textwidth}
    \centering
    \includegraphics[width=\linewidth]{
        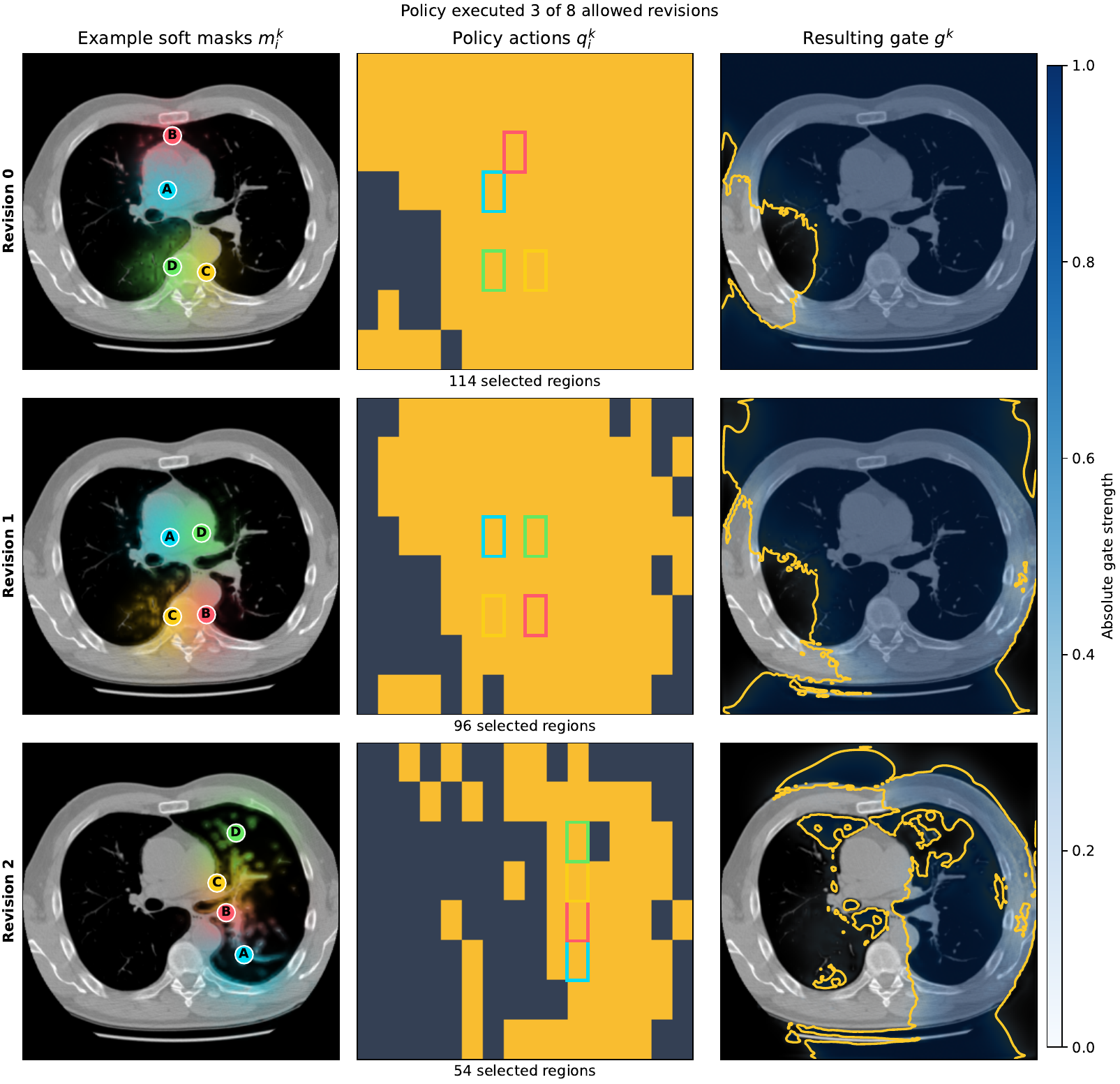}
    \caption{NLST example}
    \label{supp:fig:gate-nlst}
\end{subfigure}
\hfill
\begin{subfigure}[t]{0.49\textwidth}
    \centering
    \includegraphics[width=\linewidth]{
        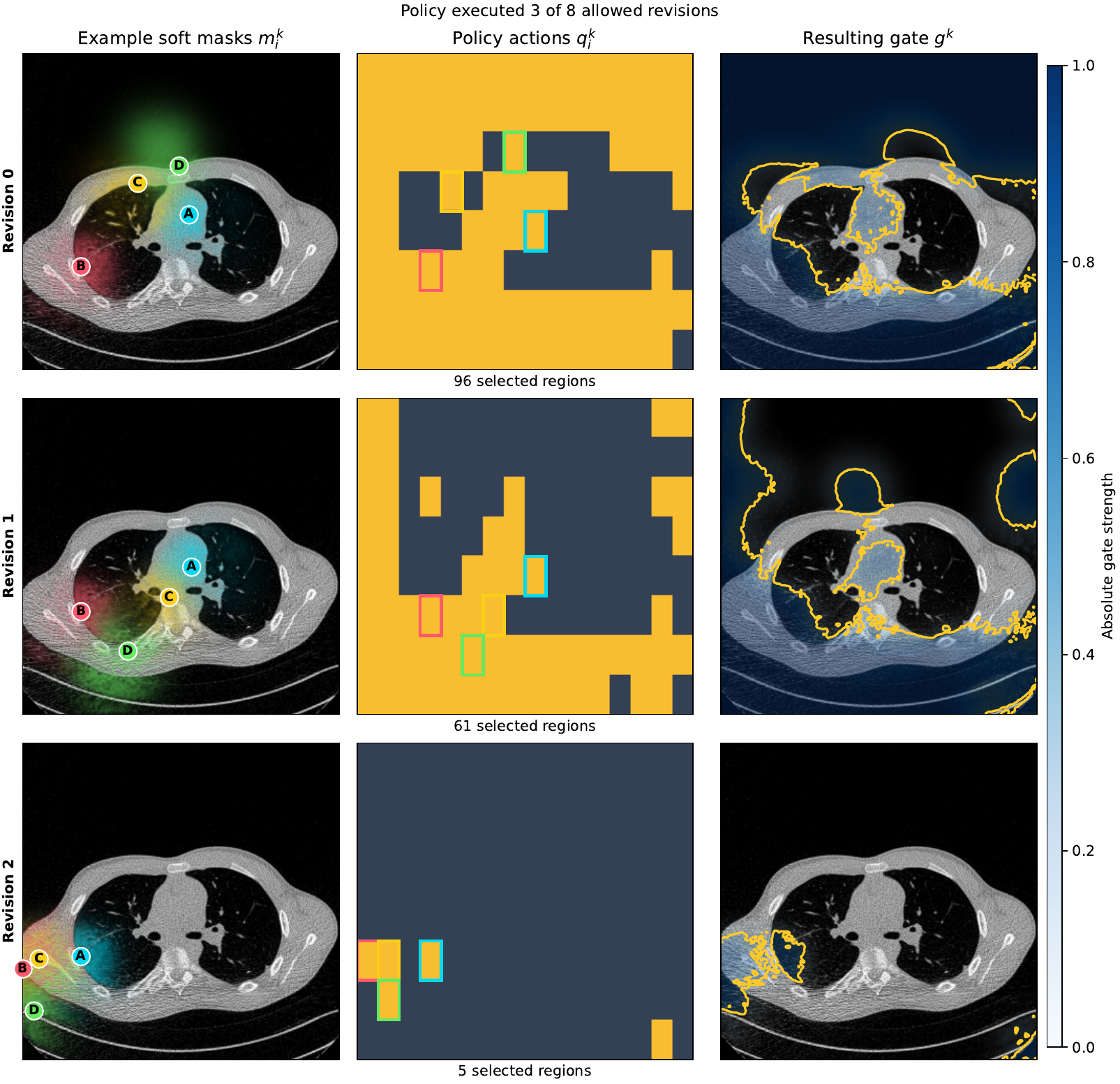}
    \caption{LDCT example}
    \label{supp:fig:gate-ldct}
\end{subfigure}

\caption{Spatial continuation behavior of IRCMF.
Each row represents an executed revision.
The left column shows four representative soft
masks, with corresponding colored outlines in
the middle policy-action grid. Yellow cells
indicate selected regions, while dark cells
indicate unselected regions. The right column
shows the resulting spatial gate formed from
all selected masks, with the $0.5$ contour
marked in yellow. Example masks are individually
rescaled for visualization, while the resulting
gate uses their original weights.}

\label{supp:fig:spatial-continuation}
\end{figure}

We examine how the learned continuation policy allocates
internal revisions across spatial regions. At revision
$k$, the router produces overlapping soft masks
$m_i^k(p)$ and binary continuation decisions $q_i^k$.
The resulting spatial gate is
\begin{equation}
\label{supp:eq:spatial-gate}
g^k(p)=\sum_i q_i^k m_i^k(p),
\qquad 0\leq g^k(p)\leq 1.
\end{equation}
The gate modulates the recurrent bottleneck update,
while the shared decoder recomputes velocity after
each executed revision. Thus, regional actions control
the spatial distribution of feature updates, whereas
early stopping reduces the number of dense decoder
evaluations.

Figures~\ref{supp:fig:gate-nlst} and
\ref{supp:fig:gate-ldct} visualize this behavior on
representative NLST and LDCT examples. Each row
corresponds to an executed revision. Four example
soft masks are displayed alongside the binary
region actions and the resulting gate. The action
grid indexes region nodes rather than image pixels,
while the gate combines all selected masks using
their original weights. The illustrated examples
execute three of the eight allowed revisions.

\medskip

\paragraph{Full-test continuation behavior.}
We evaluate the validation-selected full-model
checkpoints on all 1,695 NLST and 1,699 LDCT
test slices. Table~\ref{supp:tab:gate-analysis}
summarizes regional continuation, enhancement
quality, and observed error reductions.

\begin{table}[!htbp]
\raggedright
\normalsize
\setlength{\tabcolsep}{5pt}
\renewcommand{\arraystretch}{1.12}
\caption{Spatial continuation and enhancement
measurements on the complete NLST and LDCT test sets.
Selected-region fractions correspond to successive
revision stages. Error reductions are measured in
HU$^2$ and averaged over executed image--revision
pairs.}
\label{supp:tab:gate-analysis}
\begin{tabularx}{\linewidth}{@{}Yrr@{}}
\toprule
Measurement & NLST & LDCT\\
\midrule
Test slices & 1,695 & 1,699\\
Selected regions, revision 0 & 90.7\% & 54.7\%\\
Selected regions, revision 1 & 67.9\% & 35.4\%\\
Selected regions, revision 2 & 37.1\% & 0.9\%\\
Mean decoder evaluations & 4.42 & 3.45\\
\midrule
One-pass PSNR (dB) & 35.8408 & 32.3732\\
Learned-policy PSNR (dB) & 36.0422 & 32.3938\\
One-pass SSIM & 0.96317 & 0.84280\\
Learned-policy SSIM & 0.96388 & 0.84303\\
\midrule
Gate-weighted error reduction (HU$^2$)
& 7.651 & 4.018\\
Complement-weighted reduction (HU$^2$)
& 2.990 & 1.061\\
Revisions improving full-image error
& 89.5\% & 97.0\%\\
\bottomrule
\end{tabularx}
\end{table}

The selected-region fraction decreases from
$90.7\%$ to $37.1\%$ across the first three
revision stages on NLST and from $54.7\%$ to
$0.9\%$ on LDCT. The learned policy uses
4.42 and 3.45 decoder evaluations per slice,
respectively, compared with nine evaluations
under the complete revision schedule.
These measurements show that the continuation
policy progressively reduces regional activity
before terminating further revisions.

On NLST, $59.5\%$ of slices stop after three
revisions, $38.7\%$ after four, and $1.8\%$
after five. On LDCT, $56.7\%$ stop after two
revisions, $42.0\%$ after three, and $1.3\%$
after four. The resulting stopping distributions
are consistent with the mean decoder-evaluation
counts reported in Tab.~\ref{supp:tab:gate-analysis}.

\medskip

\paragraph{Spatial activity and enhancement.}
To examine the relationship between spatial
continuation and enhancement, we compare
gate-weighted and complement-weighted squared-error
reductions using predictions immediately before
and after each executed revision.

On NLST, the mean gate-weighted error reduction
is $7.651$~HU$^2$, compared with
$2.990$~HU$^2$ for the complement. On LDCT,
the corresponding reductions are
$4.018$ and $1.061$~HU$^2$.
Furthermore, $89.5\%$ of executed revisions
on NLST and $97.0\%$ on LDCT reduce
full-image squared error.

These results show that locations receiving
stronger spatial gating are associated with
larger observed error reductions. The comparison
characterizes the spatial behavior of the
learned policy, rather than attributing
individual pixel improvements exclusively
to the corresponding gates, since every
executed revision uses the shared dense decoder.

\medskip

% Replace these filenames with the actual
% spatial-gate figures uploaded to Overleaf.

\section{Implementation and Reproducibility Details}
\label{supp:sec:implementation}
This section specifies the configuration used in the experiments described in the main paper. The two soft-region graphs serve different purposes: the decoder graph contextualizes the current velocity prediction, whereas the bottleneck router graph aggregates information used by the next internal feature revision. The regional action gate affects the recurrent update; it does not turn a dense decoder evaluation into a sparse decoder operation.

\subsection{Model Configuration}
\label{supp:sec:model-config}

\begin{figure}[t]
\centering
\includegraphics[width=\textwidth]{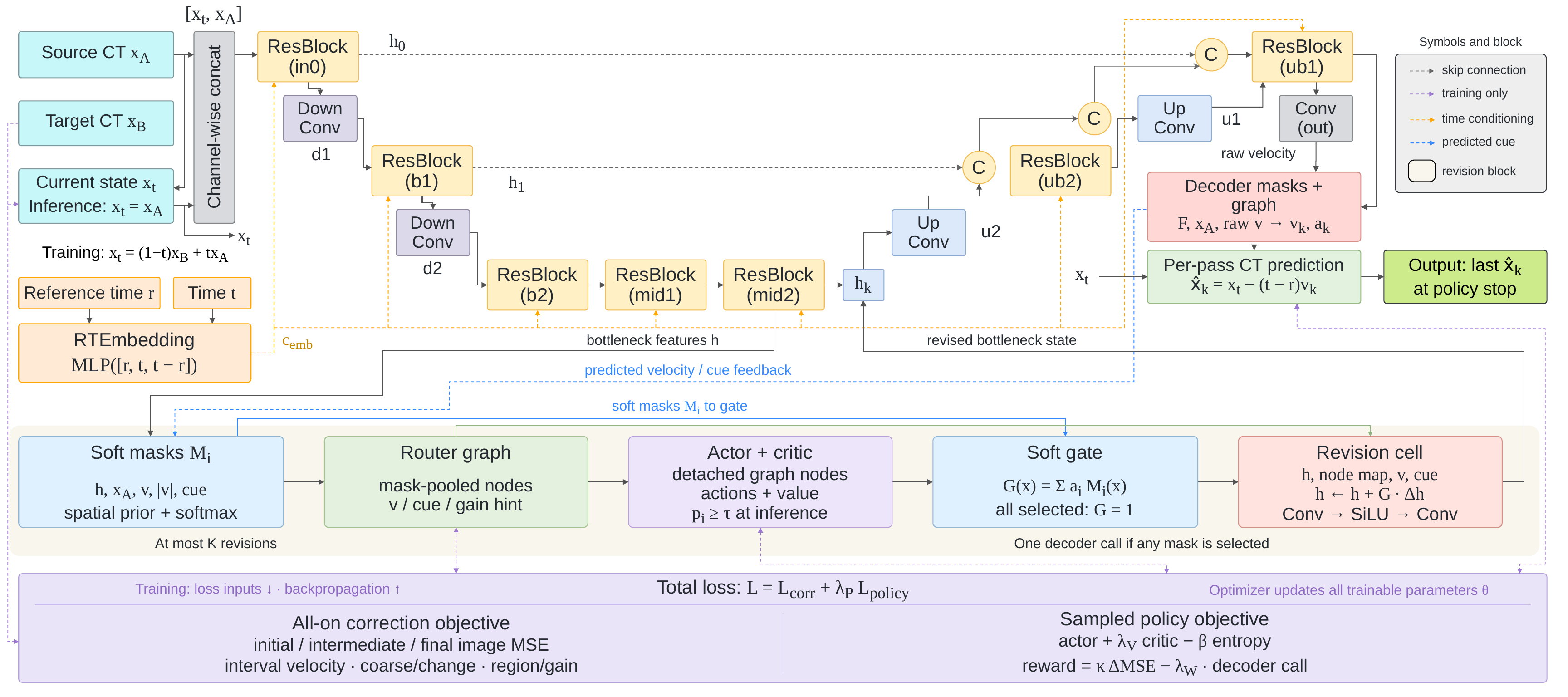}
\caption{\textbf{Detailed architecture of IRCMF.}
    The source image $x_A$ and current state $x_t$ enter a conditional U-Net.
    Its seven residual blocks share the time embedding
    $c_{\mathrm{emb}}=\mathrm{MLP}([r,t,t-r])$, where $r$ and $t$ define the
    prediction interval. At revision $k$, the shared decoder predicts a
    velocity field $v_k$ and CT image
    $\hat{x}_k=x_t-(t-r)v_k$. A decoder-level velocity-region module refines
    $v_k$; a separate bottleneck router uses the predicted velocity, its
    change cue, and image features to produce soft region masks
    $M_{k,i}(p)$, where $i$ indexes regions and $p$ is a spatial location.
    Mask-pooled features form graph nodes, and their exchanged information
    is scattered back into a spatial context map. The actor chooses binary
    actions $a_{k,i}\in\{0,1\}$, giving the revision gate
    $G_k(p)=\sum_i a_{k,i}M_{k,i}(p)$. The revision cell computes a learned
    bottleneck update $\Delta h_k$ and applies
    $h_{k+1}=h_k+G_k\odot\Delta h_k$ before reusing the same decoder.
    Any nonempty action set incurs one additional decoder evaluation.
    During training, the paired target $x_B$ constructs interval states
    and supplies image, velocity, and policy-reward supervision; inference
    uses only $x_A$ and model predictions.
    The figure uses $M_{k,i}$, $a_{k,i}$, and $G_k$ for the
    method's $m_i^k$, $q_i^k$, and $g^k$, respectively;
    subscripted states indicate the same revision indices
    as the superscripts in the method.}
\label{supp:fig:ircmf_detailed}
\end{figure}

Figure~\ref{supp:fig:ircmf_detailed} presents the
detailed architecture of IRCMF, complementing the
high-level overview in the main paper. It illustrates
the conditional U-Net structure, decoder-side regional
interaction, bottleneck router graph, and recurrent
internal revision pathway. The figure also shows how
predicted velocity and learned change cues guide
regional feature updates before the shared decoder
recomputes velocity.

The network takes the current state and the source image as two input channels and outputs one signed velocity channel. The U-Net follows a $64\!\to\!128\!\to\!256\!\to\!128\!\to\!64$ feature-width progression, with two levels of downsampling and upsampling. Time conditioning is produced by an MLP from the interval variables. Table~\ref{supp:tab:architecture} specifies the residual blocks, prediction heads, graph-node widths, revision cell, and policy settings.

\begin{table}[!htbp]
\raggedright
\normalsize
\setlength{\tabcolsep}{5pt}\renewcommand{\arraystretch}{1.12}
\caption{Backbone and revision architecture.}
\label{supp:tab:architecture}
\begin{tabularx}{\linewidth}{@{}>{\raggedright\arraybackslash}p{0.27\linewidth}Y@{}}
\toprule
Component & Configuration\\
\midrule
Input / output & Current state and source (2 channels) / signed velocity (1 channel)\\
U-Net widths & 64 $\to$ 128 $\to$ 256 $\to$ 128 $\to$ 64\\
Time embedding & MLP 3 $\to$ 256 $\to$ 256, SiLU\\
Residual blocks & 7; two 3$\times$3 convolutions each; GroupNorm (8 groups), SiLU\\
Down / up sampling & Two levels; 4$\times$4 convolutions, stride 2, padding 1; skip concatenation\\
Prediction heads & 3$\times$3 convolutions for velocity and change cues\\
Graph node widths & Decoder: 32; router: 64\\
Revision cell & 514 $\to$ 256 $\to$ 256; 3$\times$3 convolutions, SiLU, \texttt{tanh} output\\
Revision embedding & 8 $\times$ 64\\
Revision scale $\gamma$ & 0.1\\
Policy threshold $\tau$ & 0.5\\
\bottomrule
\end{tabularx}
\end{table}

The shared recurrent reviser takes bottleneck features, graph context, and velocity-related cues as inputs. Its update is scaled by $\gamma=0.1$; the inference policy uses a probability threshold of $\tau=0.5$. The revision-step embedding has eight positions, consistent with a maximum of eight additional decoder evaluations after the initial prediction. The complete execution therefore uses at most nine dense decoder calls per slice.

Table~\ref{supp:tab:graph} gives the spatial and normalization settings for the two graph sites at $512\times512$ input resolution. Both use an $8\times8$ anchor lattice, but the router has two candidate masks per anchor, while the decoder graph uses one. The mask resolutions and spatial priors consequently differ; these quantities should not be interpreted as equivalent spatial partitions.

\begin{table}[!htbp]
\raggedright
\normalsize
\setlength{\tabcolsep}{5pt}\renewcommand{\arraystretch}{1.12}
\caption{Regional assignment and graph settings at 512$\times$512 input resolution.}
\label{supp:tab:graph}
\begin{tabularx}{\linewidth}{@{}>{\raggedright\arraybackslash}p{0.42\linewidth}YY@{}}
\toprule
Setting & Decoder & Router\\
\midrule
Anchor grid & 8 $\times$ 8 & 8 $\times$ 8\\
Mask resolution & 64 $\times$ 64 & 128 $\times$ 128\\
Spatial prior & $-1.5d^2$ & $-128d^2$\\
Cue SD floor & 0.05 & 0.02\\
Squared-gap cap & 16 & None\\
Affinity initialization & $\log 2$ & $\log 2$\\
Assignment temperature & 1 & 1\\
\bottomrule
\end{tabularx}
\par\smallskip\raggedright
Decoder assignments use feature-similarity weight 2, support $d^2\le4.5$, and 8-neighbor graph connectivity. Router cues are clipped to $[-4,4]$. Each router anchor has two candidates with speed prior $-0.5(s-c)^2$ and centers $c=\pm0.75$. The decoder-feedback offset is 0.5.
\end{table}

For the decoder graph, feature similarity has weight 2, spatial support is limited to $d^2\le4.5$, and graph communication uses the immediate eight-neighbor lattice connections. Router cue values are clipped to $[-4,4]$. Its two speed-prior centers are $c=\pm0.75$, with the reported prior $-0.5(s-c)^2$. The decoder feedback uses an offset of 0.5. Affinity coefficients are initialized to an effective value of $\log2$ at both graph sites. These numerical settings describe the reported implementation; they are not independently tuned results.

\subsection{Training and Evaluation Protocols}
\label{supp:sec:protocols}
\paragraph{Optimization and supervision.}
Table~\ref{supp:tab:training} records the optimization schedule, batch accumulation, sampling ranges, objective weights, and policy cost. The configuration runs for 30 epochs with a constant learning-rate schedule, weight decay $10^{-4}$, seed 1729, and a global gradient-norm clipping threshold of 1.0. During training, the finite interval is sampled with $t\sim U(0.05,1)$ and $r\sim U(0,t)$.

\begin{table}[!htbp]
\raggedright
\normalsize
\setlength{\tabcolsep}{6pt}
\renewcommand{\arraystretch}{1.12}
\caption{Common training and evaluation protocol
for all methods on the NLST dataset.}
\label{supp:tab:common-protocol}
\begin{tabularx}{\linewidth}{@{}
>{\raggedright\arraybackslash}p{0.27\linewidth}Y@{}}
\toprule
Protocol Element & Settings\\
\midrule
Image preparation &
$512\times512$ CT slices; $[-800,600]$ HU window;
target aligned to input\\

Data partitions &
15,095 training slices (29 patients);
3,579 validation slices (7 patients)\\

Training duration &
30 epochs; best checkpoint selected using
validation PSNR\\

Optimization &
AdamW; learning rate $10^{-4}$ for the model
and $3\times10^{-4}$ for policy components,
where applicable; effective batch size 32
across two GPUs\\

Evaluation &
Selected checkpoint evaluated on 1,695 slices
from three held-out patients; full-image HU
PSNR and SSIM averaged across slices\\
\bottomrule
\end{tabularx}
\end{table}

All comparison methods follow the common training
and evaluation protocol in
Table~\ref{supp:tab:common-protocol}.
The same patient-disjoint partitions, image
preparation, training duration, and evaluation
procedure are used to ensure consistency across
methods. The additional learning rate for policy
components applies only to methods that include
a learned policy. Architecture-specific objectives
and components remain method-dependent.

\begin{table}[!htbp]
\raggedright
\normalsize
\setlength{\tabcolsep}{5pt}\renewcommand{\arraystretch}{1.12}
\caption{Optimization and loss weights for IRCMF.}
\label{supp:tab:training}
\begin{tabularx}{\linewidth}{@{}>{\raggedright\arraybackslash}p{0.62\linewidth}Y@{}}
\toprule
Setting & Value\\
\midrule
Weight decay & $10^{-4}$\\
Schedule / epochs & Constant / 30\\
Microbatch / accumulation & 4 images per GPU / 4 steps\\
Seed / gradient-norm clip & 1729 / 1.0\\
Interval sampling & $t\sim U(0.05,1)$; $r\sim U(0,t)$\\
Correction weights (final, initial, intermediate, regional) & 1, 0.3, 0.2, 0.1\\
Auxiliary loss weights (interval, coarse, cue, gain) & 0.25, 0.05, 0.005, 0.01\\
Policy weights (critic, entropy, overall) & 0.5, 0.001, 1\\
Gain and reward scale $\kappa$ & 200\\
Decoder-call cost & 0.002\\
Advantage clipping & $[-2,2]$\\
\bottomrule
\end{tabularx}
\end{table}

The supervised objective weights the final, initial, intermediate, and regional terms by $1$, $0.3$, $0.2$, and $0.1$. The interval, coarse, cue, and gain auxiliary terms use weights $0.25$, $0.05$, $0.005$, and $0.01$. The gain-target and reward scale is $\kappa=200$; gain targets are clipped to $[-2,2]$, and the Smooth-L1 transition is 1. The reported architecture controls use final, initial, and interval supervision weights $1$, $0.3$, and $0.25$.

The actor--critic objective uses critic weight $0.5$ and entropy weight $0.001$. A penalty of $0.002$ is charged when an action leads to an additional dense decoder call. The policy advantage is clipped to $[-2,2]$. These settings define the reported training configuration; the region actions are not treated as independent measurements of per-region causal benefit.

\paragraph{DICOM preprocessing and pairing.}
The images are processed as individual two-dimensional slices, without adjacent-slice context or training augmentation. DICOM stored pixel values are converted to Hounsfield units (HU) using the reported rescale slope and intercept. Declared padding pixels are set to $-800$~HU. Intensities are then clipped to $[-800,600]$~HU and linearly mapped to $[-1,1]$.

The paired images are aligned using DICOM metadata translations and bilinear resampling with border padding. Spatial resizing to the model input resolution also uses bilinear interpolation, with \texttt{align\_corners=False}. As a consequence, all reported image-level and regional measurements are tied to the specified alignment and HU-window protocol.

\paragraph{Inference.}
Inference follows Algorithm~1 of the main paper. The initial dense decoder produces a velocity estimate before any internal revision. At each subsequent stage, still-eligible router regions can trigger a feature update. Once no region is selected, no further dense decoder evaluation is performed; regions that have stopped remain inactive. The decoder-evaluation count includes the initial prediction and every executed revision. It does not count regional node updates as independent decoder calls.

\paragraph{Evaluation.}
Evaluation uses FP32 and reports PSNR and SSIM against the paired target after conversion to the analysis HU range. Both metrics use a data range of 1400~HU. SSIM uses an $11\times11$ Gaussian window with standard deviation 1.5 and constants $K_1=0.01$ and $K_2=0.03$. These details are relevant when reproducing the absolute scores or comparing models that may otherwise use a different intensity range or SSIM implementation.

\subsection{Dataset Splits and Subject Identifiers}
\label{supp:sec:dataset-splits}
The full-image NLST and LDCT evaluations use their respective patient-disjoint training, validation, and test partitions. Table~\ref{supp:tab:subjects} records the subject identifiers used in each partition. The separate NLSTseg subset supplies the five cases for tumor-ROI and radiomic analysis; it is not presented as an additional part of either full-image test split. Each reported paired sample consists of a source slice and its corresponding aligned target slice.

\begin{table}[!htbp]
\raggedright
\normalsize
\setlength{\tabcolsep}{5pt}\renewcommand{\arraystretch}{1.16}
\caption{Patient identifiers for the NLST and LDCT-and-Projection-data splits and the separate NLSTseg analysis subset.}
\label{supp:tab:subjects}
\begin{tabularx}{\linewidth}{@{}>{\raggedright\arraybackslash}p{0.20\linewidth}Y@{}}
\toprule
Analysis Subset & Subject IDs\\
\midrule
\multicolumn{2}{@{}l}{\textbf{NLST Subset}}\\
\midrule
Train & 200044, 200328, 200378, 200667, 200718, 200933, 200979, 201317, 201377, 201510, 201627, 202095, 202232, 202853, 203007, 203208, 203828, 204203, 204382, 204502, 204469, 204603, 204612, 204847, 204855, 204862, 204934, 205419, 205769\\
Validation & 202249, 203067, 203357, 203681, 203951, 205659, 205786\\
Test & 200087, 200480, 201233\\
\midrule
\addlinespace[3pt]
\multicolumn{2}{@{}l}{\textbf{LDCT-and-Projection-data}}\\
\midrule
Train & C002, C012, C016, C021, C067, C077, C099, C107, C120, C121, C124, C128, C130, C135, C158, C160, C162, C166, C170, C179, C190, C193, C202, C218, C219, C227, C232, C234, C241, C246, C249, C252, C257, C258, C267, C268, C280, C295, C296\\
Validation & C004, C052, C111, C203, C224\\
Test & C027, C050, C081, C095, C261\\
\addlinespace[3pt]
\midrule
\multicolumn{2}{@{}l}{\textbf{NLSTseg Subset}}\\
\midrule
Tumor ROI and radiomics & 200209, 201519, 202610, 203769, 205687\\
\bottomrule
\end{tabularx}
\end{table}

\subsection{Model Size and Training Configuration}
\label{supp:sec:parameters}
\label{supp:sec:coverage}
Parameter counts provide a complementary description of model size alongside decoder-evaluation count and latency. Table~\ref{supp:tab:parameters} lists the recorded total and trainable parameter counts.

\begin{table}[!htbp]
\raggedright
\normalsize
\setlength{\tabcolsep}{6pt}\renewcommand{\arraystretch}{1.07}
\caption{Parameter counts of the evaluated CT models.}
\label{supp:tab:parameters}
\begin{tabularx}{\linewidth}{@{}>{\raggedright\arraybackslash}p{0.50\linewidth}>{\raggedleft\arraybackslash}p{0.22\linewidth}>{\raggedleft\arraybackslash}X@{}}
\toprule
Method & Total Parameters & Trainable Parameters\\
\midrule
CNN-baseline & 148,929 & 148,929\\
STAN-CT & 25,185,924 & 25,183,172\\
RadiomicGAN & 45,265,024 & 45,265,024\\
CMF & 6,147,841 & 6,147,841\\
ACL & 4,567,772 & 4,567,772\\
MaIR-S & 1,417,345 & 1,417,345\\
MambaIRv2 & 22,651,705 & 22,651,705\\
RA-CMF & 6,236,075 & 6,236,075\\
IRCMF & 8,067,407 & 8,066,351\\
CLIPDenoising & 34,515,043 & 10,987,779\\
G2CR-FPN & 126,509,253 & 120,242,373\\
ASCON & 4,914,230 & 4,914,230\\
\bottomrule
\end{tabularx}
\end{table}

% \medskip

% IRCMF contains $8{,}067{,}407$ total parameters, of which $8{,}066{,}351$ are trainable. Table~\ref{supp:tab:parameter-breakdown} attributes the stored parameters to the CMF U-Net backbone, decoder-region graph, cue heads, router and policy components, and shared bottleneck revision cell. \textcolor{red}{The full count includes 1,056 frozen parameters in an inherited unused projection.} The table describes parameter storage, not the computational work of any particular inference schedule.

\begin{table}[!htbp]
\raggedright
\normalsize
\setlength{\tabcolsep}{6pt}\renewcommand{\arraystretch}{1.07}
\caption{IRCMF parameter breakdown.}
\label{supp:tab:parameter-breakdown}
\begin{tabularx}{\linewidth}{@{}>{\raggedright\arraybackslash}p{0.71\linewidth}>{\raggedleft\arraybackslash}X@{}}
\toprule
Component & Parameters\\
\midrule
Conditional CMF U-Net backbone & 6,147,841\\
Decoder soft-region graph and velocity feedback & 35,142\\
Bottleneck coarse velocity and change heads & 4,610\\
Decoder change-cue head & 577\\
Soft-region router, graph, actor, and critic & 104,645\\
Shared bottleneck revision cell & 1,774,592\\
\midrule
Total & 8,067,407\\
\bottomrule
\end{tabularx}
\end{table}

Table~\ref{supp:tab:coverage} describes the recorded split consistency and training provenance for the evaluated methods. ``Same split'' refers to the task-specific experimental partition, not to a single shared split across the two datasets.

\FloatBarrier

Table~\ref{supp:tab:coverage} describes the recorded
split consistency and training provenance for the evaluated
methods. ``Same split'' refers to the task-specific
experimental partition, not to a single shared split
across the two datasets.

\begin{table}[H]
\raggedright
\normalsize
\setlength{\tabcolsep}{3pt}
\renewcommand{\arraystretch}{1.05}
\caption{Training provenance and data-split consistency
for the evaluated methods, within each dataset.}
\label{supp:tab:coverage}
\begin{tabularx}{\linewidth}{@{}>{\raggedright\arraybackslash}p{0.42\linewidth}>{\raggedright\arraybackslash}p{0.24\linewidth}>{\raggedright\arraybackslash}X@{}}
\toprule
Method & Data Split Configuration & Trained From Scratch\\
\midrule
CNN-baseline & Same split & Yes\\
STAN-CT & Same split & Yes\\
RadiomicGAN & Same split & Yes\\
CMF & Same split & Yes\\
ACL & Same split & Yes\\
MaIR-S & Same split & Yes\\
MambaIRv2 & Same split & Yes\\
RA-CMF & Same split & Yes\\
CLIPDenoising & Same split & Yes$^\dagger$\\
G2CR-FPN & Same split & Yes\\
ASCON & Same split & Yes\\
IRCMF (Ours) & Same split & Yes\\
\bottomrule
\end{tabularx}
\par\smallskip
\noindent$^\dagger$CLIPDenoising's denoiser is trained from
scratch with a normal-dose-only objective; its CLIP RN50
encoder is initialized from pretrained weights and kept frozen.
\end{table}

\section{Class-Wise Radiomic Feature Concordance}
\label{supp:sec:radiomics}
The main paper reports overall radiomic concordance and tumor-region image fidelity. Here, we provide the more detailed breakdown across six radiomic feature groups on the same five-patient NLSTseg analysis subset with 71 tumor-positive paired slices. The groups are first-order statistics, gray-level co-occurrence matrix (GLCM), gray-level run-length matrix (GLRLM), gray-level size-zone matrix (GLSZM), gray-level dependence matrix (GLDM), and neighboring gray-tone difference matrix (NGTDM). Values are presented as mean $\pm$ standard deviation of the reported feature-level concordance correlation coefficients (CCC) within each class.
\begin{table}[!htbp]
\raggedright
\normalsize
\setlength{\tabcolsep}{3.0pt}\renewcommand{\arraystretch}{1.15}
\caption{Class-wise radiomic feature concordance on the five-patient NLSTseg subset. Entries are mean $\pm$ SD across feature-level CCCs within each class. OP, AR, and LP denote one pass, all revisions, and learned policy; * marks our model. Bold and underline indicate the highest and second-highest distinct displayed means; ties share markings.}
\label{supp:tab:radiomics}
\begin{tabular*}{\textwidth}{@{\extracolsep{\fill}}lrrrrrr@{}}
\toprule
Method & First-order $\uparrow$ & GLCM $\uparrow$ & GLRLM $\uparrow$ & GLSZM $\uparrow$ & GLDM $\uparrow$ & NGTDM $\uparrow$ \\
\midrule
Input & $0.59\pm0.20$ & $0.45\pm0.27$ & $0.54\pm0.25$ & $0.54\pm0.25$ & $0.52\pm0.21$ & $0.40\pm0.26$ \\
CNN-baseline & $0.79\pm0.11$ & $0.55\pm0.28$ & $0.79\pm0.14$ & $0.79\pm0.14$ & $0.79\pm0.12$ & $0.48\pm0.25$ \\
STAN-CT & $0.88\pm0.14$ & $0.83\pm0.20$ & $0.65\pm0.29$ & $0.65\pm0.29$ & $0.64\pm0.22$ & $0.80\pm0.25$ \\
RadiomicGAN & $0.91\pm0.05$ & $0.89\pm0.13$ & $\boldsymbol{0.91\pm0.06}$ & $\boldsymbol{0.91\pm0.06}$ & $\boldsymbol{0.94\pm0.04}$ & $0.90\pm0.05$ \\
CMF & $0.97\pm0.05$ & $0.91\pm0.14$ & $0.83\pm0.20$ & $0.83\pm0.20$ & $0.86\pm0.12$ & $0.90\pm0.09$ \\
RA-CMF & $0.96\pm0.06$ & $0.93\pm0.11$ & $\boldsymbol{0.91\pm0.09}$ & $\boldsymbol{0.91\pm0.09}$ & $\boldsymbol{0.94\pm0.05}$ & \underline{$0.96\pm0.03$} \\
ACL & \underline{$0.98\pm0.03$} & $0.93\pm0.14$ & $0.84\pm0.19$ & $0.84\pm0.19$ & $0.90\pm0.09$ & $0.94\pm0.03$ \\
MaIR-S & \underline{$0.98\pm0.03$} & \underline{$0.96\pm0.04$} & $\boldsymbol{0.91\pm0.07}$ & $\boldsymbol{0.91\pm0.07}$ & $0.92\pm0.05$ & $0.92\pm0.05$ \\
MambaIRv2 & \underline{$0.98\pm0.02$} & $0.95\pm0.07$ & $0.83\pm0.21$ & $0.83\pm0.21$ & $0.89\pm0.10$ & $0.95\pm0.03$ \\
\midrule
IRCMF (OP)* & $\boldsymbol{0.99\pm0.01}$ & $\boldsymbol{0.97\pm0.03}$ & $\boldsymbol{0.91\pm0.09}$ & $\boldsymbol{0.91\pm0.09}$ & $\boldsymbol{0.94\pm0.05}$ & $0.95\pm0.03$ \\
IRCMF (AR)* & $\boldsymbol{0.99\pm0.02}$ & $0.94\pm0.18$ & \underline{$0.90\pm0.10$} & \underline{$0.90\pm0.10$} & \underline{$0.93\pm0.06$} & $0.94\pm0.05$ \\
IRCMF (LP)* & $\boldsymbol{0.99\pm0.01}$ & $0.93\pm0.19$ & $0.89\pm0.12$ & $0.89\pm0.12$ & $0.92\pm0.07$ & $\boldsymbol{0.97\pm0.01}$ \\
\bottomrule
\end{tabular*}

\end{table}

\section{Additional Qualitative Comparisons}
\label{supp:sec:qualitative}

\subsection{NLST Reconstruction-Kernel Enhancement}
\label{supp:sec:qual-nlst}

\begingroup
\centering
\captionof{figure}{Five qualitative examples of
reconstruction-kernel enhancement on NLST.
The STANDARD input and predictions from the
comparison methods are shown against the paired
LUNG reference. Highlighted regions are enlarged
for visual comparison. PSNR and SSIM correspond
to each displayed slice.}
\label{supp:fig:nlst_examples}
\par\medskip

\begin{overpic}[width=\textwidth]{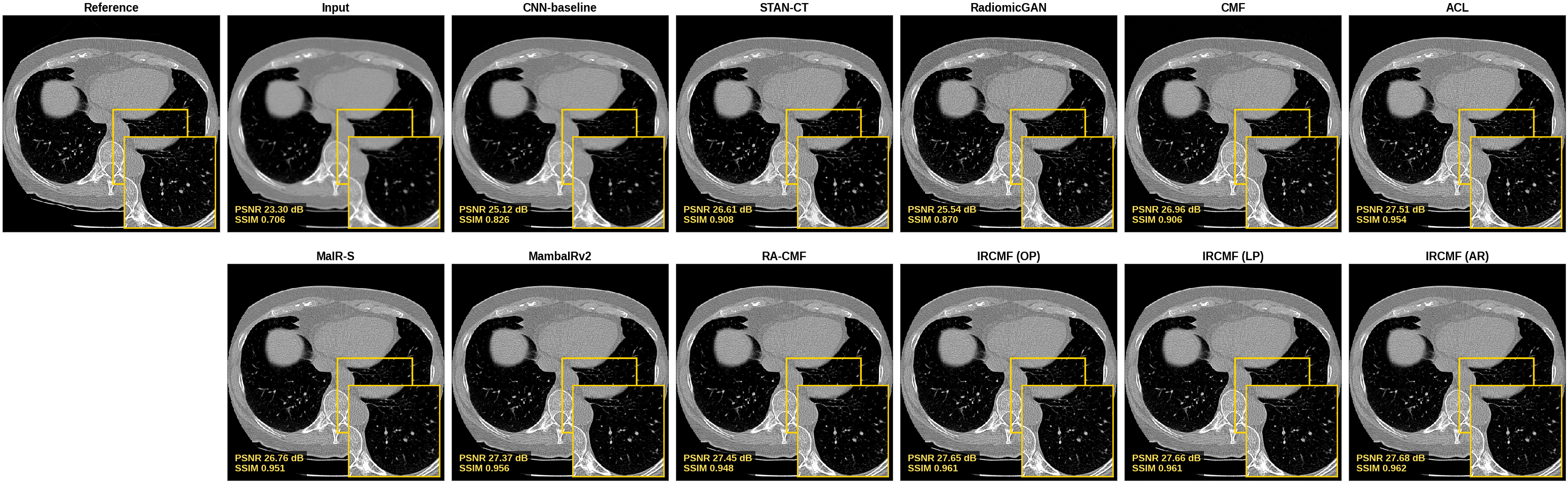}
    \put(1.5,2){\large\bfseries (a)}
\end{overpic}
\par\medskip

\begin{overpic}[width=\textwidth]{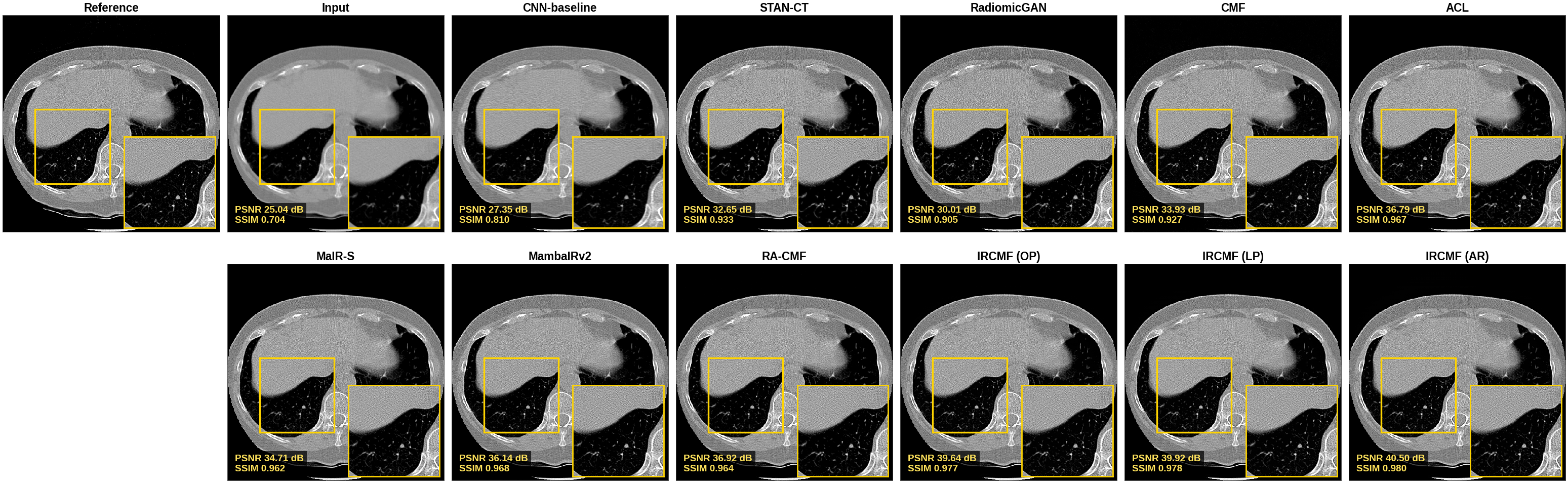}
    \put(1.5,2){\large\bfseries (b)}
\end{overpic}
\par\medskip

\begin{overpic}[width=\textwidth]{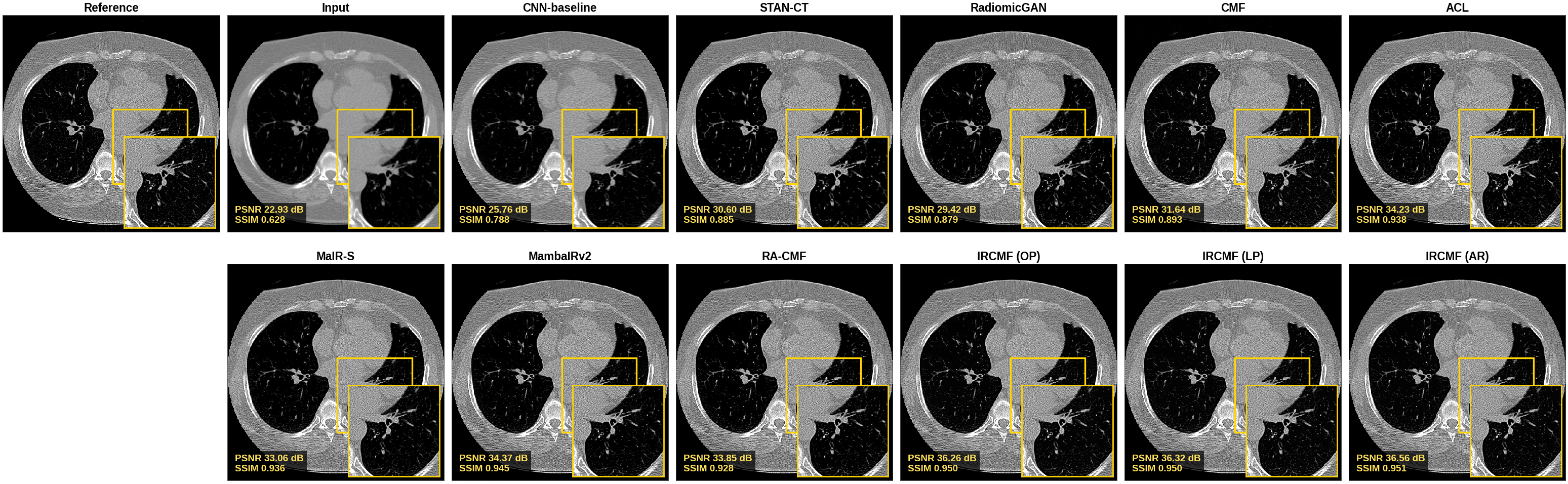}
    \put(1.5,2){\large\bfseries (c)}
\end{overpic}
\par\medskip

\begin{overpic}[width=\textwidth]{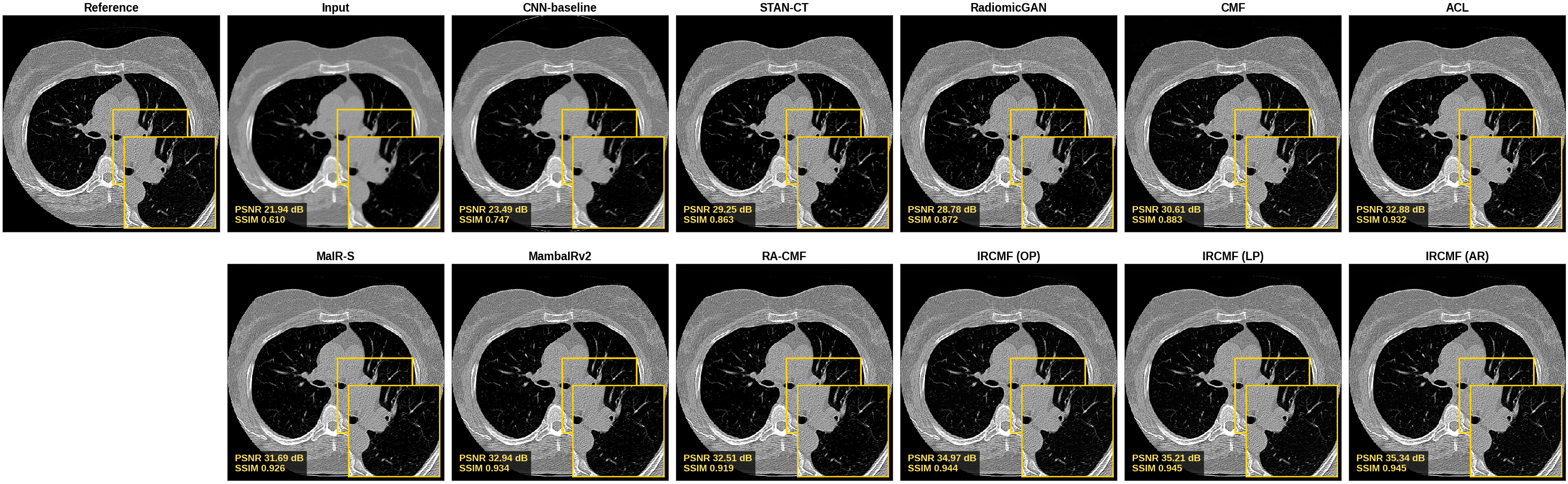}
    \put(1.5,2){\large\bfseries (d)}
\end{overpic}
\par\medskip

\begin{overpic}[width=\textwidth]{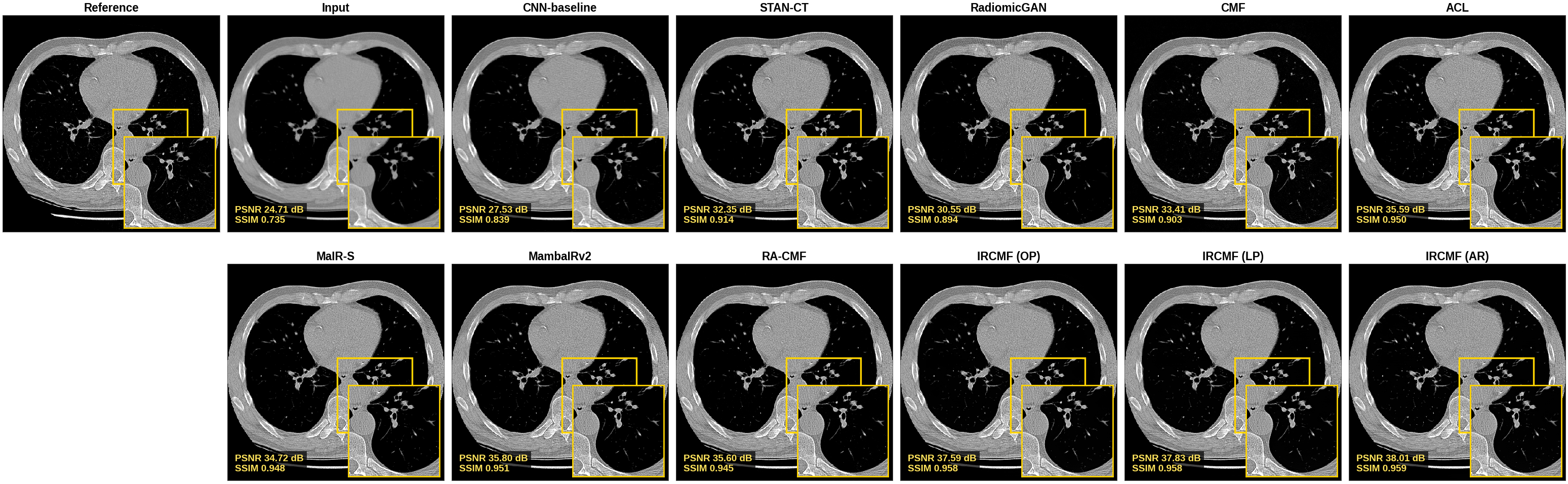}
    \put(1.5,2){\large\bfseries (e)}
\end{overpic}
\par
\endgroup

\clearpage
\subsection{LDCT Dose Enhancement}
\label{supp:sec:qual-ldct}

\begingroup
\centering
\captionof{figure}{Five qualitative examples of
low-to-routine-dose enhancement on
LDCT-and-Projection-data. The input, paired
reference, baseline predictions, and IRCMF
outputs under one-pass (OP), learned-policy (LP),
and all-revisions (AR) inference are compared.
Highlighted regions are enlarged for visual
comparison. PSNR and SSIM correspond to each
displayed slice.}
\label{supp:fig:ldct_examples}
\par\medskip

\begin{overpic}[width=\textwidth]{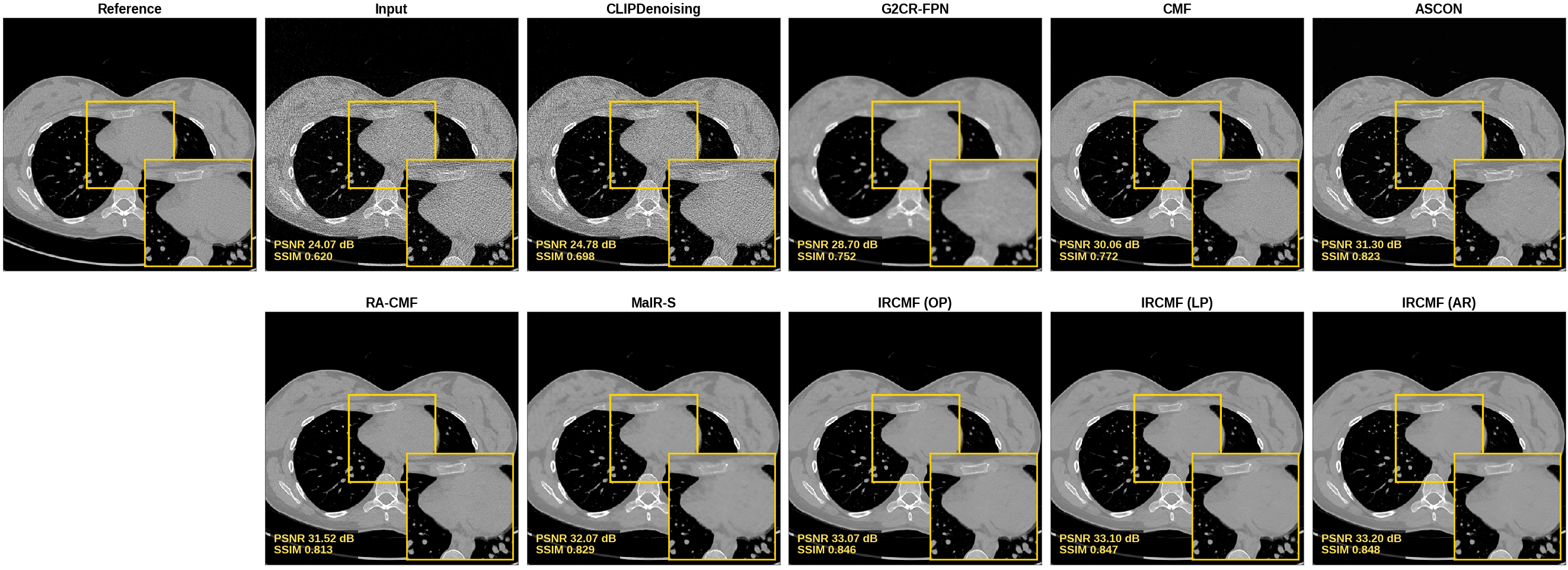}
    \put(1.5,2){\large\bfseries (a)}
\end{overpic}
\par\medskip

\begin{overpic}[width=\textwidth]{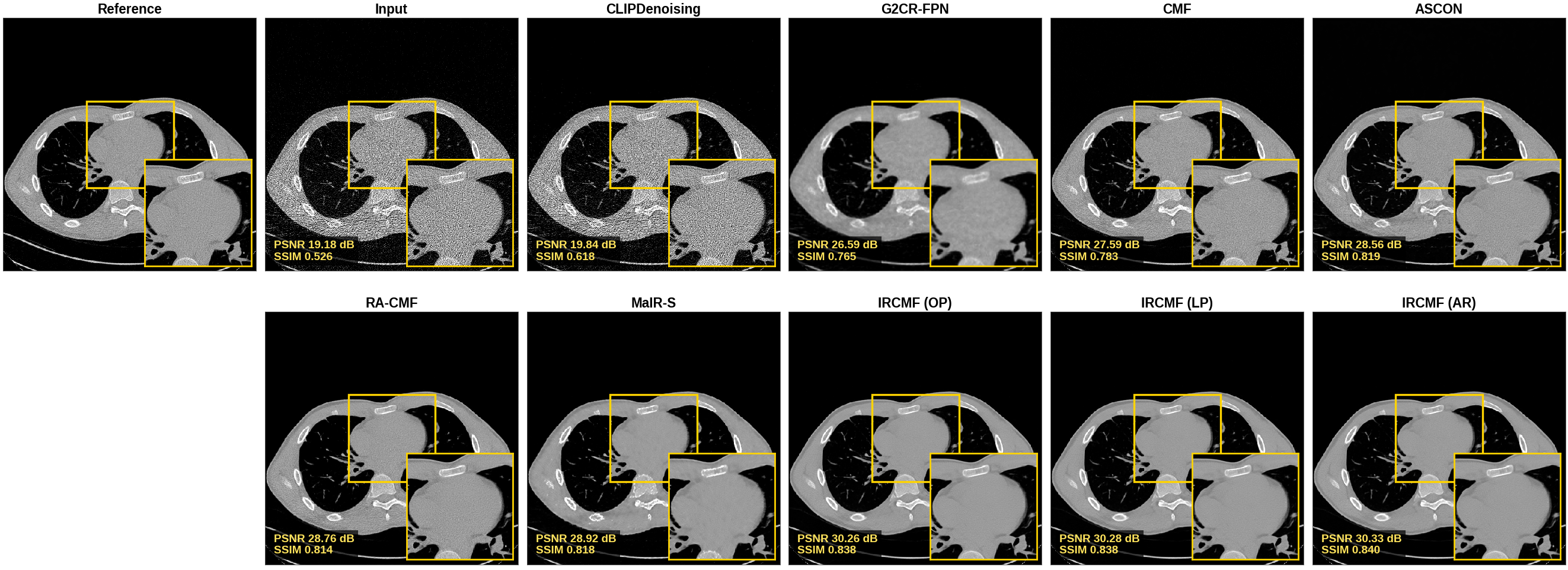}
    \put(1.5,2){\large\bfseries (b)}
\end{overpic}
\par\medskip

\begin{overpic}[width=\textwidth]{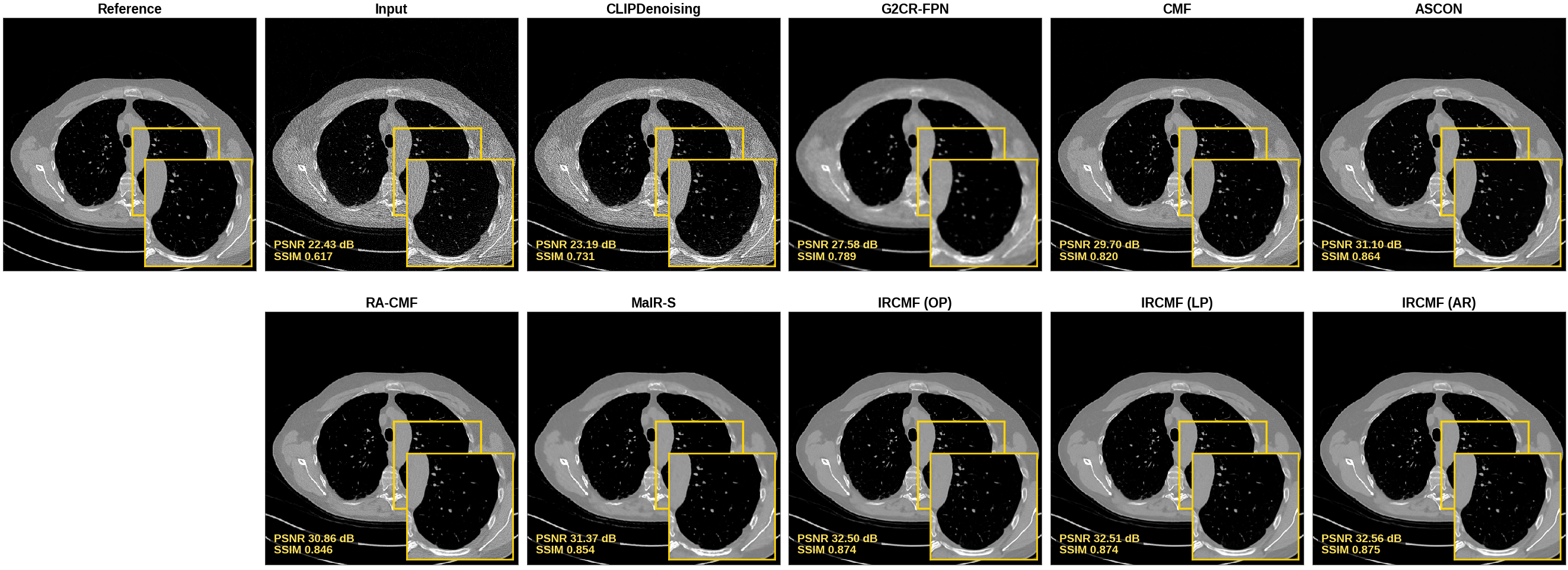}
    \put(1.5,2){\large\bfseries (c)}
\end{overpic}
\par\medskip

\begin{overpic}[width=\textwidth]{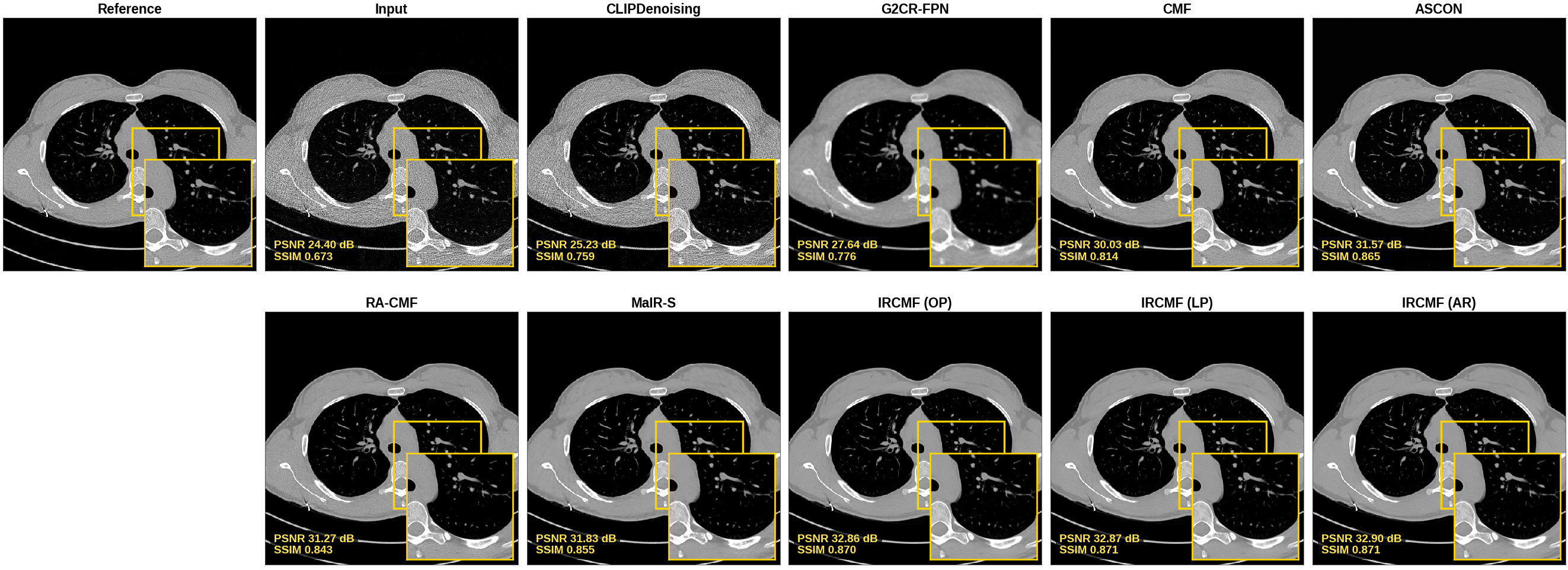}
    \put(1.5,2){\large\bfseries (d)}
\end{overpic}
\par\medskip

\begin{overpic}[width=\textwidth]{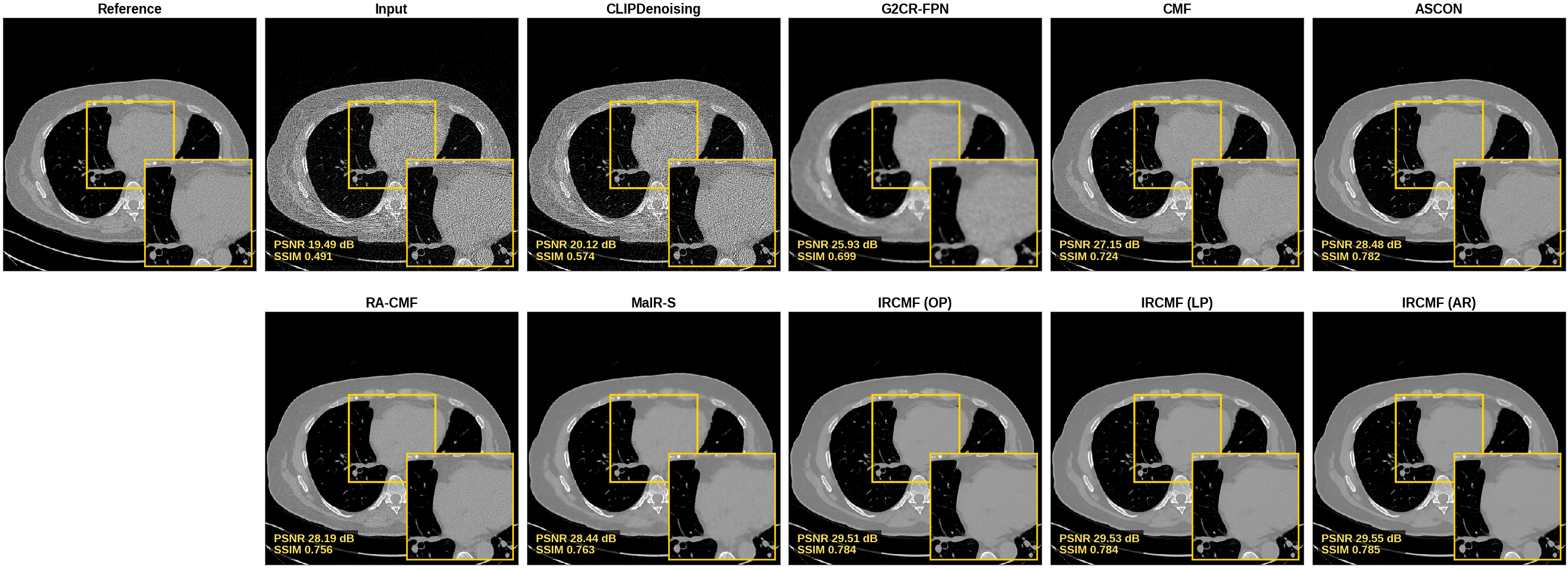}
    \put(1.5,2){\large\bfseries (e)}
\end{overpic}
\par

\endgroup

\clearpage
\subsection{NLSTseg Tumor-Region Enhancement}
\label{supp:sec:qual-nlstseg}

\begingroup
\centering
\captionof{figure}{Five qualitative examples of
tumor-region enhancement on the NLSTseg subset.
The input, paired reference, baseline predictions,
and IRCMF inference modes are compared.
Highlighted regions provide enlarged views
of tumor-containing areas.}
\label{supp:fig:nlstseg_examples}
\par\medskip

\begin{overpic}[width=\textwidth]{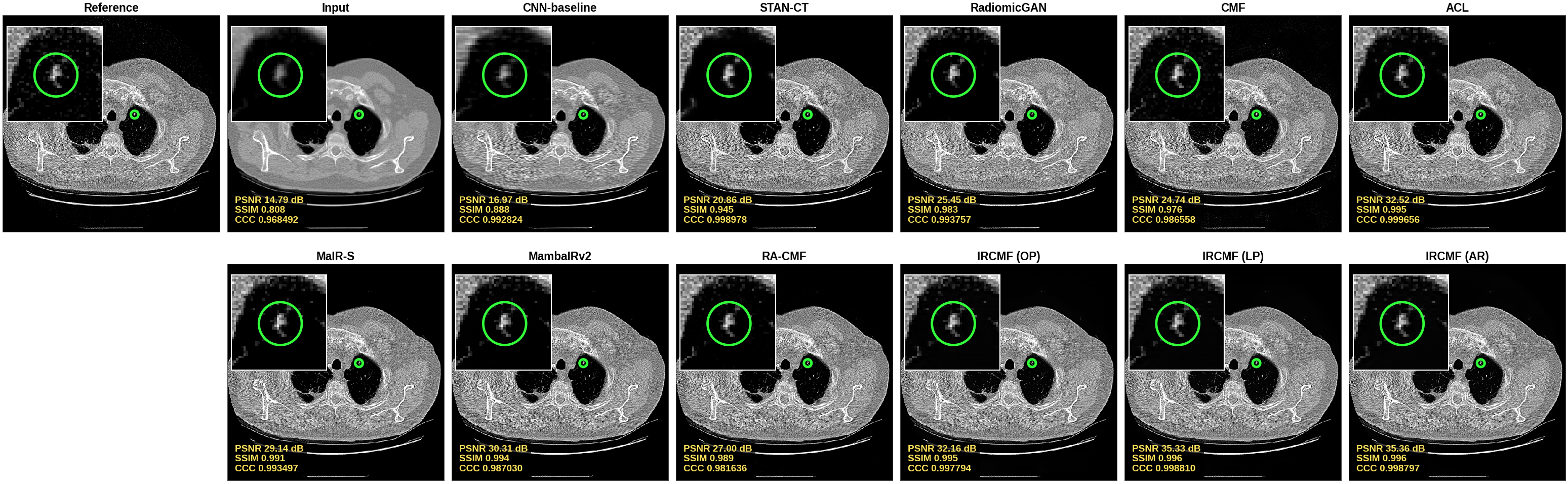}
    \put(1.5,2){\large\bfseries (a)}
\end{overpic}
\par\medskip

\begin{overpic}[width=\textwidth]{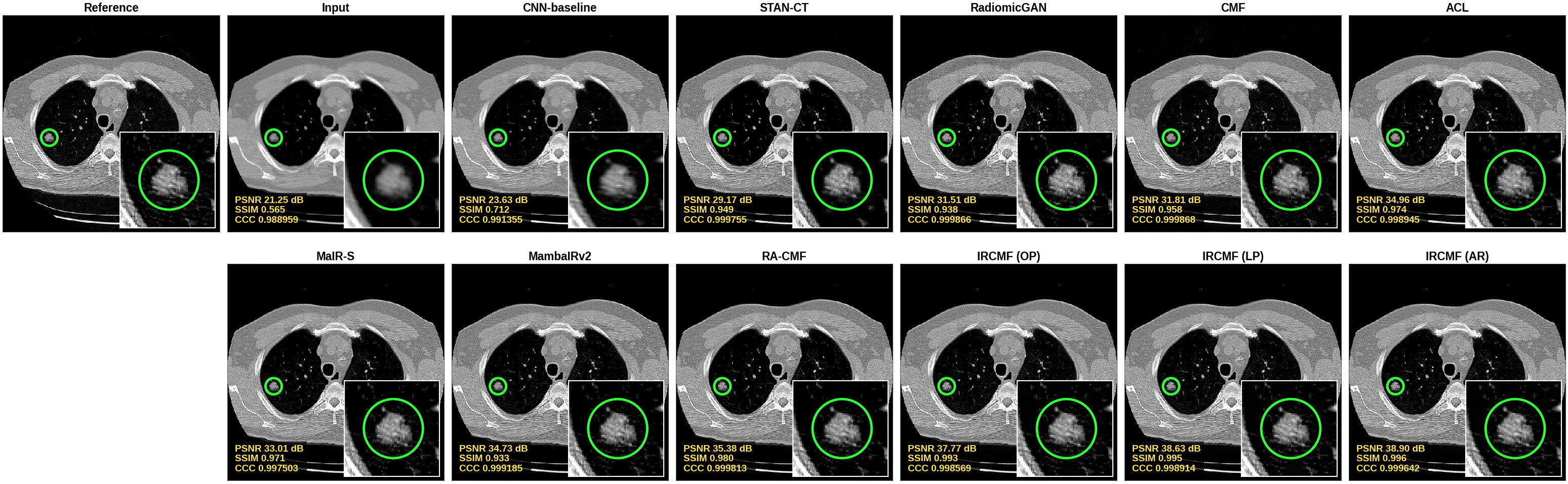}
    \put(1.5,2){\large\bfseries (b)}
\end{overpic}
\par\medskip

\begin{overpic}[width=\textwidth]{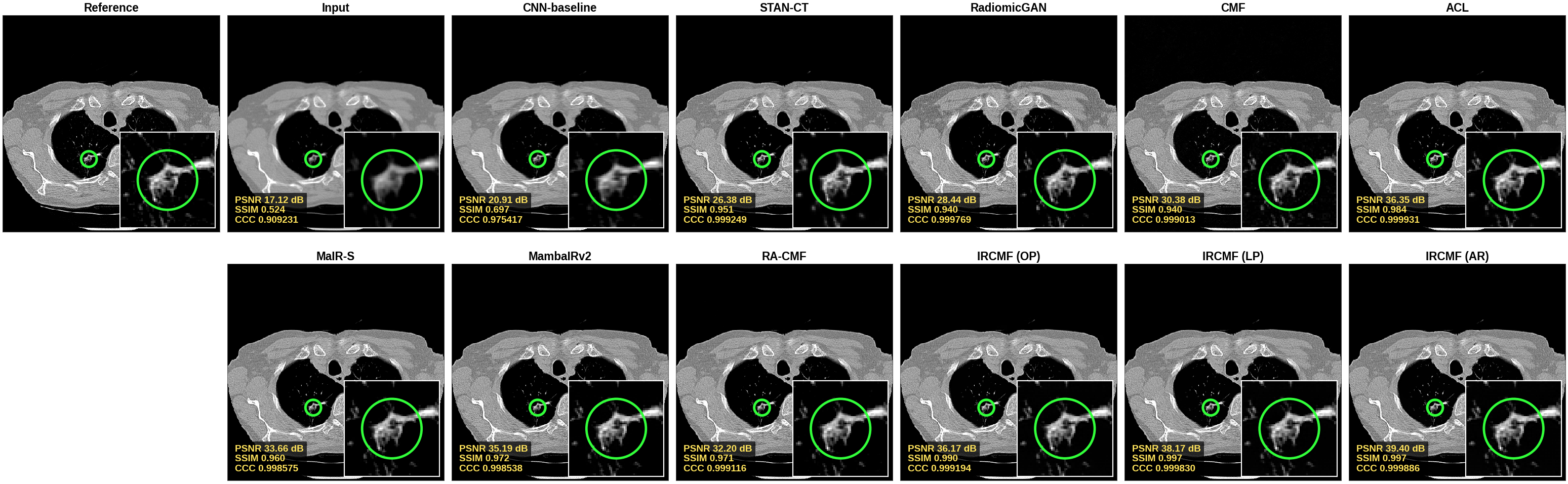}
    \put(1.5,2){\large\bfseries (c)}
\end{overpic}
\par\medskip

\begin{overpic}[width=\textwidth]{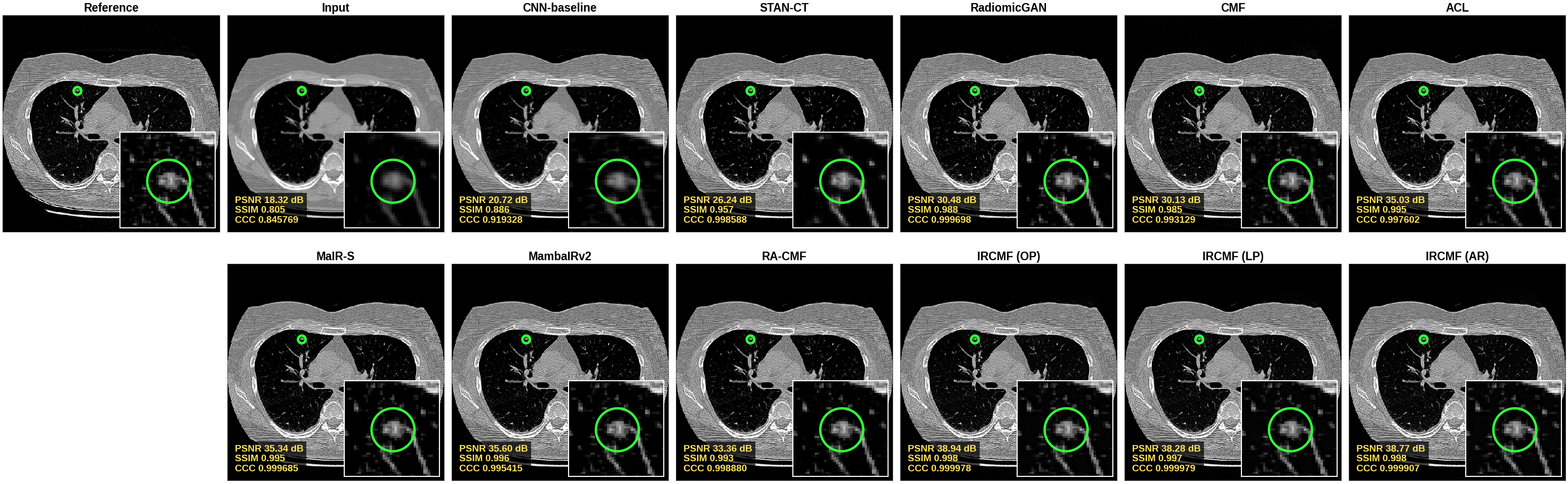}
    \put(1.5,2){\large\bfseries (d)}
\end{overpic}
\par\medskip

\begin{overpic}[width=\textwidth]{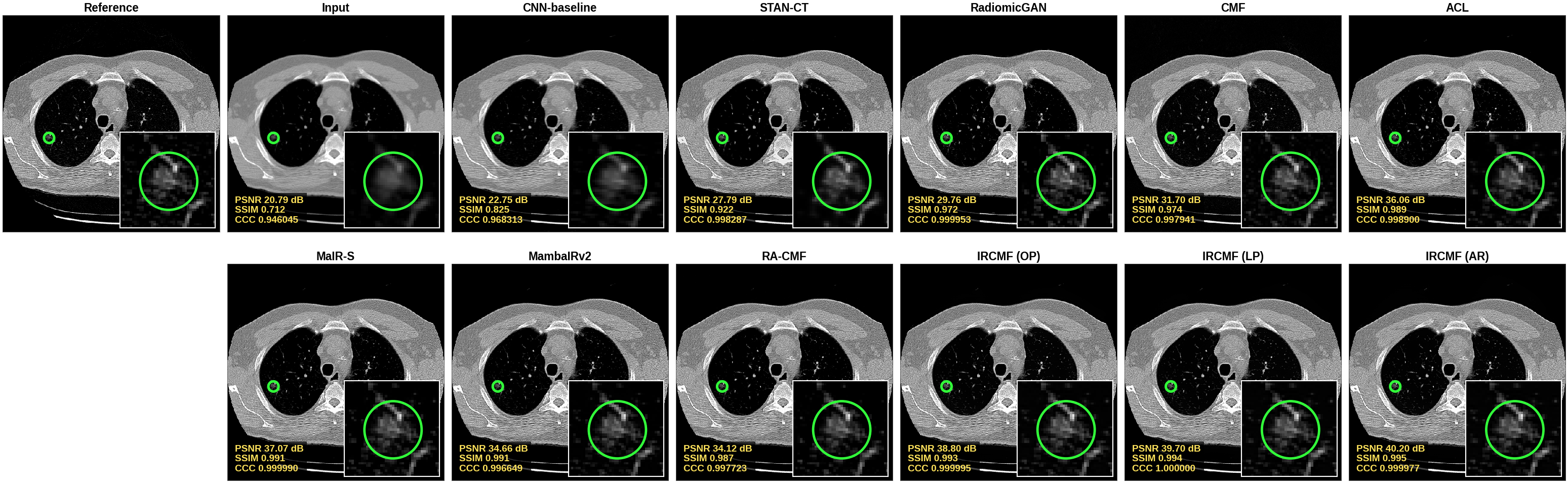}
    \put(1.5,2){\large\bfseries (e)}
\end{overpic}
\par
\endgroup